%% file: main.tex
\documentclass{article}

\PassOptionsToPackage{numbers, compress}{natbib}
\usepackage[numbers]{natbib}

\newif\ifpreprint

\preprintfalse  

\ifpreprint
  \usepackage[preprint]{neurips_2025}
\else
  \usepackage[final]{neurips_2025}
\fi

\usepackage[utf8]{inputenc} 
\usepackage[T1]{fontenc}    
\usepackage{hyperref}       
\usepackage{url}            
\usepackage{booktabs}       
\usepackage{amsfonts}       
\usepackage{nicefrac}       
\usepackage{microtype}      
\usepackage[table]{xcolor}         
\usepackage[nolist]{acronym}
\usepackage{amsmath}
\usepackage{float}
\usepackage{tabularx}
\usepackage{graphicx}
\usepackage{subcaption}

\usepackage{hyperref}
\usepackage{cleveref}
\usepackage{wrapfig}

\title{Diffusion-Based Hierarchical Graph Neural Networks for Simulating Nonlinear Solid Mechanics}

\input{03_definitions/authors}
\input{03_definitions/commands}

\begin{document}
\input{03_definitions/acronyms}

\maketitle

\input{00_sections/00_abstract}

\input{00_sections/01_introduction}

\input{00_sections/02_related_work}

\input{00_sections/03_method}

\input{00_sections/04_experiments}

\input{00_sections/05_results}

\input{00_sections/06_conclusion}
\ifpreprint
\else
\input{00_sections/07_acknowledgements}
\fi

\bibliographystyle{unsrtnat}
\bibliography{04_bibliography/bib}


\newpage
\appendix
\input{02_appendix/0X_broader_impact}

\input{02_appendix/0X_datasets}

\input{02_appendix/0X_setup}

\input{02_appendix/0X_baselines}
\input{02_appendix/0X_results}

\ifpreprint
\else
    \clearpage
    \newpage
    \input{02_appendix/0X_neurips_checklist}
\fi




\end{document}

%% file: 03_definitions/authors.tex
\author{\textbf{Tobias Würth}$^1$\thanks{correspondence to \texttt{tobias.wuerth@kit.edu}}~~
\textbf{Niklas Freymuth}$^2$~
\textbf{Gerhard Neumann}$^2$~
\textbf{Luise Kärger}$^1$~
\\
$^1$Institute of Vehicle System Technology, 
Karlsruhe Institute of Technology, Karlsruhe\\
$^2$Autonomous Learning Robots,
Karlsruhe Institute of Technology,
Karlsruhe\\
}



%% file: 03_definitions/commands.tex
\definecolor{oldtextcolor}{RGB}{111, 123, 140}     
\definecolor{llmtextcolor}{RGB}{94, 129, 172}      
\definecolor{wiptextcolor}{RGB}{208, 135, 112}     
\definecolor{todoitemcolor}{RGB}{191, 97, 106}     
\definecolor{commenttextcolor}{RGB}{163, 190, 140} 



%% file: 03_definitions/acronyms.tex
\begin{acronym}[ABCDEF] 
\acro{gns}[GNS]{Graph Network Simulator}
\acro{mgn}[MGN]{\textsc{MeshGraphNet}}
\acroplural{mgn}[MGNs]{\textsc{MeshGraphNets}}
\acro{pimgn}[PI-MGN]{Physics-Informed \textsc{MeshGraphNet}}
\acroplural{pimgn}[PI-MGNs]{Physics-Informed \textsc{MeshGraphNets}}
\acro{gnn}[GNN]{Graph Neural Network}
\acro{hgnn}[HGNN]{Hierarchical Graph Neural Network}
\acro{mpn}[MPN]{Message Passing Network}
\acro{mlp}[MLP]{Multilayer Perceptron}
\acro{cnn}[CNN]{Convolutional Neural Network}
\acro{mse}[MSE]{Mean Squared Error}
\acro{pde}[PDE]{Partial Differential Equation}
\acro{gmm}[GMM]{Gaussian Mixture Model}
\acro{fem}[FEM]{Finite Element Method}
\acro{fvm}[FVM]{Finite Volume Method}
\acro{fdm}[FDM]{Finite Difference Method}
\acro{ml}[ML]{Machine Learning}
\acro{amg}[AMG]{Algebraic multigrid}
\acro{ddpm}[DDPM]{Denoising Diffusion Probabilistic Model}
\acro{umgn}[UMGN]{U-hierarchical~\textsc{MeshGraphNet}}
\acroplural{umgn}[UMGNs]{U-hierarchical~\textsc{MeshGraphNets}}
\acro{ampn}[AMPN]{Algebraic-hierarchical Message Passing Network}
\acro{unet}[U-Net]{U-Net}
\acro{robin}[ROBIN]{Rolling Diffusion-Batched Inference Network}
\acro{robi}[ROBI]{Rolling Diffusion-Batched Inference}
\acro{silu}[SiLU]{Sigmoid Linear Unit}
\acro{pderef}[PDE-Refiner]{PDERef}
\acro{hcmt}[HCMT]{Hierarchical Contact Mesh Transformer}
\acro{amr}[AMR]{Adaptive Mesh Refinement}
\acro{bsdl}[BSDL]{Bi-stride-coarsening and Delaunay-remeshing}
\acro{intramps}[Intra-MP-Stack]{Intra-Level-Message Passing Stack}
\acro{intermps}[Inter-MP-Stack]{Inter-Level-Message Passing Stack}
\acro{cmt}[CMT]{Contact Mesh Transformer}
\acro{hmt}[HMT]{Hierarchical Mesh Transformer}
\acro{bsms}[BSMS]{Bi-Stride Multi-Scale GNN}

\acro{abl_10_ds}[10 diffusion steps]{10 diffusion steps}
\acro{abl_5_ds}[5 diffusion steps]{5 diffusion steps}
\acro{abl_wo_shared}[w/o shared layer]{w/o shared layer}
\acro{abl_wo_diffusion}[w/o diffusion]{w/o diffusion}
\acro{abl_wo_hierarchy}[w/o hierarchy]{w/o hierarchy}
\acro{abl_state}[State prediction]{State prediction}
\acro{abl_hcmt}[HCMT model]{HCMT model}


\end{acronym}

%% file: 00_sections/00_abstract.tex
\begin{abstract}

Graph-based learned simulators have emerged as a promising approach for simulating physical systems on unstructured meshes, offering speed and generalization across diverse geometries. 
However, they often struggle with capturing global phenomena, such as bending or long-range correlations usually occurring in solid mechanics, and suffer from error accumulation over long rollouts due to their reliance on local message passing and direct next-step prediction.
We address these limitations by introducing the Rolling Diffusion-Batched Inference Network (ROBIN), a novel learned simulator that integrates two key innovations: (i) Rolling Diffusion-Batched Inference (ROBI), a parallelized inference scheme that amortizes the cost of diffusion-based refinement across physical time steps by overlapping denoising steps across a temporal window. (ii) A Hierarchical Graph Neural Network built on algebraic multigrid coarsening, enabling multiscale message passing across different mesh resolutions. This architecture, implemented via Algebraic-hierarchical Message Passing Networks, captures both fine-scale local dynamics and global structural effects critical for phenomena like beam bending or multi-body contact. 
We validate ROBIN on challenging $2$D and $3$D solid mechanics benchmarks involving geometric, material, and contact nonlinearities. ROBIN achieves state-of-the-art accuracy on all tasks, substantially outperforming existing next-step learned simulators while reducing inference time by up to an order of magnitude compared to standard diffusion simulators.

\end{abstract}

%% file: 00_sections/01_introduction.tex
\vspace{-0.1cm}
\section{Introduction}
\vspace{-0.2cm}

Physical simulations enable many engineering and scientific fields to gain quick insights into complex systems or to evaluate design decisions.  
Conventional simulations model the physical system using \acp{pde}.
Usually, the \ac{pde} is discretized by numerical methods, such as the~\ac{fem}~\citep{larsonFiniteElementMethod2013}, the~\ac{fvm}~\citep{moukalledFiniteVolumeMethod2016a}, or the~\ac{fdm}~\citep{ozisikFiniteDifferenceMethods2017}.
This process reduces the need for cumbersome, resource-intensive real-world experiments.
Recent research aims to speed up simulation with~\ac{ml}-based models~\citep{sanchez-gonzalezLearningSimulateComplex2020,pfaffLearningMeshBasedSimulation2020}.
These learned simulators promise to allow researchers and practitioners to evaluate large amounts of virtual, simulated experiments.
These simulations in turn unlock applications in engineering design and manufacturing optimization~\citep{allenInverseDesignFluidStructure2022,zimmerlingOptimisationManufacturingProcess2022a,wurthPhysicsinformedNeuralNetworks2023}.

This work aims to improve learned simulators, focusing on simulations of nonlinear solid mechanics as a representative class of examples.
We combine recent image-based~\acp{ddpm}~\citep{lippePDERefinerAchievingAccurate2023,ruheRollingDiffusionModels2024,kohlBenchmarkingAutoregressiveConditional2024b} with~\acp{hgnn}~\citep{liuMultiresolutionGraphNeural2021,fortunatoMultiScaleMeshGraphNets2022,linoMultiscaleRotationequivariantGraph2022a} (cf. \Cref{fig:hierarchical_graph_diffusion}).
While diffusion has shown promising results on images~\citep{
dhariwal2021diffusion, ho2020denoising, rombach2022high}, audio~\citep{
kong2020diffwave,chen2020wavegrad} and even policy learning for robotics~\citep{chi2023diffusion, scheikl2024movement}, it suffers from cost-intensive inference due to its iterative denoising procedure.
We alleviate this high inference cost on time-dependent domains with~\ac{robi}, a novel scheduling scheme that batches denoising steps of consecutive time steps.
\ac{robi} already starts denoising future prediction steps by using partially refined previous steps.
    This process reduces the number of model evaluations to the number of time steps and preserves the time-shift equivariance of Markovian systems.
\ac{robi} only affects the inference process, allowing us to utilize conventional, parallelized~\ac{ddpm} training.

\input{05_inputs/figures/hierarchical_graph_diffusion}

We combine~\ac{robi} with~\acp{ddpm} and~\acfp{ampn} to form the~\acf{robin}, which significantly accelerates simulation while improving predictive accuracy.
\ac{robin} constitutes the first diffusion-based refiner for simulating physical dynamics on unstructured meshes, surpassing the current accuracy ceiling of~\acp{hgnn}.
We train~\acp{robin} on three challenging $2$D and $3$D solid mechanics datasets involving geometric, material, and contact nonlinearities. 
Across all datasets,~\ac{robin} significantly improves over state-of-the-art mesh-based simulators~\citep{pfaffLearningMeshBasedSimulation2020, yuLearningFlexibleBody2023} in terms of predictive accuracy. 
Leveraging~\acp{ddpm},~\ac{robin} accurately captures low-frequency global solution modes while resolving high-frequency components.
We further find that our proposed inference method,~\ac{robi}, speeds up diffusion-based inference for learned simulations by up to an order of magnitude while maintaining accuracy.
In addition, the~\ac{ampn} architecture of~\ac{robin} enables efficient transfer to much larger meshes, while maintaining near-\ac{fem} accuracy.\footnote{Project page, code and datasets are available at https://tbswrth.github.io/ROBIN.}

To summarize, we
i) propose \ac{robi}, a novel inference scheduling scheme for diffusion-based simulators that amortizes denoising across time steps, reducing inference to a single model evaluation per step while preserving time-shift equivariance;  
ii) introduce \ac{robin}, a diffusion-based \ac{hgnn} for nonlinear solid mechanics that combines multiscale message passing with \ac{robi} to provide fast, accurate diffusion-based simulations;  
iii) demonstrate state-of-the-art performance on challenging $2$D and $3$D benchmarks, outperforming existing simulators in both accuracy and runtime.

%% file: 05_inputs/figures/hierarchical_graph_diffusion.tex
\begin{figure}[!htb]
    \centering
    \includegraphics[width=\textwidth,trim={0pt 220pt 0pt 170pt},  clip,]{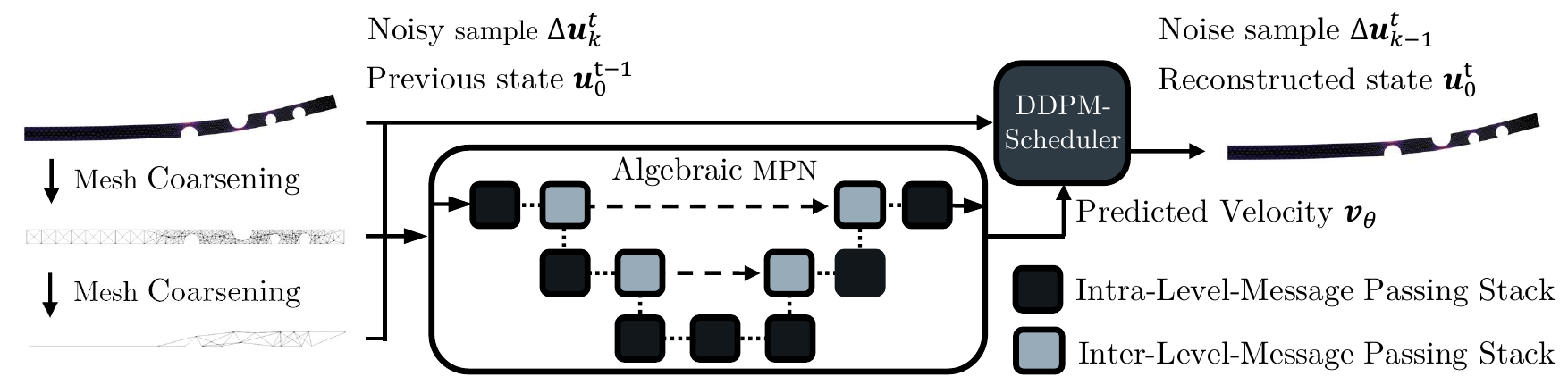}
    \caption{
        Overview of a~\ac{robin} prediction.~\ac{robin} coarsens the fine mesh multiple times with~\ac{amg} to create a graph hierarchy.~\ac{robin} predicts the denoising velocity $\mathbf{v}_\theta(\Delta \mathbf{u}^t_k, k, \mathbf{u}^{t-1}_0)$ at time step $t$ using~\acp{ampn}, given a noisy residual sample $\Delta {\mathbf{u}}^{t}_{k}$, the diffusion step $k$, and a previous state $\mathbf{u}^{t-1}_{0}$. The prediction is used to draw a new noisy residual sample $\Delta {\mathbf{u}}^{t}_{k-1}$ and to update the state $\mathbf{u}^{t}_{0}$.
    }
    \label{fig:hierarchical_graph_diffusion}
    \vspace{-0.2cm}
\end{figure}

%% file: 00_sections/02_related_work.tex
\vspace{-0.1cm}
\section{Related Work}
\vspace{-0.2cm}

\textbf{Simulating Complex Physics.}
Simulating complex physical systems often requires numerical solvers, such as the~\ac{fem}~\citep{reddyIntroductionFiniteElement2005,brennerMathematicalTheoryFinite2008,larsonFiniteElementMethod2013}, the~\ac{fvm}~\citep{moukalledFiniteVolumeMethod2016a}, or the~\ac{fdm}~\citep{ozisikFiniteDifferenceMethods2017}. 
While accurate, numerical solvers scale poorly with problem complexity, often requiring multiple hours or even days for a single rollout on a modern workstation. 
Recent work shows that~\ac{ml}-based models are able to learn such numerical simulations from data~\citep{guoConvolutionalNeuralNetworks2016,longPDENetLearningPDEs2018, sanchez-gonzalezLearningSimulateComplex2020, pfaffLearningMeshBasedSimulation2020, brandstetterMessagePassingNeural2021, lippePDERefinerAchievingAccurate2023, yuLearningFlexibleBody2023, ruheRollingDiffusionModels2024}.
\ac{ml}-based models provide speed-ups of one to two orders of magnitude while being fully differentiable, which accelerates downstream applications such as design~\citep{allenInverseDesignFluidStructure2022} or manufacturing process optimization~\citep{zimmerlingOptimisationManufacturingProcess2022a, wurthPhysicsinformedNeuralNetworks2023}. 
Many learned simulators operate autoregressively.
They mimic numerical solvers by using their own predictions to estimate the residual between successive time steps~\citep{sanchez-gonzalezLearningSimulateComplex2020, pfaffLearningMeshBasedSimulation2020, linkerhagner2023grounding}.
Similarly, Neural ODEs~\citep{chenNeuralOrdinaryDifferential2018, iakovlevLearningContinuoustimePDEs2020} predict time derivatives and advance solutions via numerical integration.
In contrast to these supervised approaches, Physics-Informed Neural Networks~\citep{raissiPhysicsinformedNeuralNetworks2019} directly operate on a~\ac{pde} loss function to train~\acp{mlp}~\citep{amininiakiPhysicsinformedNeuralNetwork2021,sunSurrogateModelingFluid2020,wurthPhysicsinformedNeuralNetworks2023},~\acp{cnn}~\citep{gaoPhyGeoNetPhysicsinformedGeometryadaptive2021,shuPhysicsinformedDiffusionModel2023} or~\acp{gnn}~\citep{gaoPhysicsinformedGraphNeural2022, wurthPhysicsinformedMeshGraphNetsPIMGNs2024a}.
Finally, Neural Operators aim to learn mesh-independent solution operators~\citep{liFourierNeuralOperator2020, wenUFNOAnEnhancedFourier2022, luLearningNonlinearOperators2021, brandstetterMessagePassingNeural2021, haoGNOTGeneralNeural}.
\ac{robin} is an autoregressive learned simulator that replaces direct next-step prediction with multiple denoising diffusion steps, leveraging generative inference to improve prediction accuracy.

\textbf{Learning Mesh-based Simulations with \acf{gnn}.}
\citet{pfaffLearningMeshBasedSimulation2020} introduced \textsc{MeshGraphNets}, a \ac{mpn}-based simulator that encodes simulation states as graphs using mesh connectivity and physical proximity. 
While accurate for small problems, \acp{mgn} does not scale well, as its receptive field is limited by the number of message-passing steps in the \ac{mpn}. 
To address this issue, recent work expands the receptive field via global attention~\citep{jannyEAGLELargescaleLearning2022, hanPredictingPhysicsMeshreduced2022b, alkinUniversalPhysicsTransformers2024} or hierarchical mesh representations using Graph Convolutional Networks~\citep{montiGeometricDeepLearning2017} and~\acp{mpn}~\citep{battagliaRelationalInductiveBiases2018a, fortunatoMultiScaleMeshGraphNets2022}. 
Extensions further improve efficiency and accuracy through rotation equivariance~\citep{linoMultiscaleRotationequivariantGraph2022a}, hierarchical edge design~\citep{grigorevHOODHierarchicalGraphs2023}, bi-stride pooling~\citep{caoEfficientLearningMeshBased2023}, and attention mechanisms acting on edges and across hierarchies~\citep{yuLearningFlexibleBody2023}.
\citep{pereraMultiscaleGraphNeural2024b} leverages~\ac{amr} to create mesh hierarchies for multi-scale~\acp{gnn}.
Complementary,~\aclp{pimgn}~\citep{wurthPhysicsinformedMeshGraphNetsPIMGNs2024a} integrate FEM-based training to improve accuracy and robustness.
Existing methods are generally trained using a next-step \ac{mse} loss, which favors learning lower solution frequencies at the cost of accuracy in higher frequency bands that have less impact on the loss~\citep{lippePDERefinerAchievingAccurate2023}.
However, autoregressive models trained with a \ac{mse} loss often overlook low-amplitude frequencies~\citep{lippePDERefinerAchievingAccurate2023}.
Our approach is orthogonal, coupling hierarchical~\ac{gnn} with denoising diffusion models.
This approach takes advantage of the large receptive field of multi-scale \acp{gnn} while pushing accuracy toward diffusion limits.

\textbf{Diffusion-based Simulations.}
Diffusion models have been applied to physics-informed image super-resolution~\citep{shuPhysicsinformedDiffusionModel2023}, flow field reconstruction~\citep{liMultiScaleReconstructionTurbulent2023}, and steady-state flow generation on grids using \acp{cnn}~\citep{lienenZeroTurbulenceGenerative2023}, and more recently to meshes with hierarchical \acp{gnn}~\citep{valenciaLearningDistributionsComplex2024}. 
These models, however, operate on isolated frames and do not capture time-dependent dynamics. 
In contrast, our model predicts deterministic physical evolution rather than equilibrium samples via autoregressive rollouts.

While next-step simulators trained with \ac{mse} loss capture high-amplitude, low-frequency components, they often miss low-amplitude components~\citep{lippePDERefinerAchievingAccurate2023}. 
\acp{pderef} address this via iterative refinement, improving long-horizon accuracy of grid-based \ac{cnn} simulators. 
We extend this idea to unstructured meshes by combining algebraic mesh coarsening~\citep{vanekAlgebraicMultigridSmoothed1996b, ruge4AlgebraicMultigrid1987} with hierarchical \acp{gnn}~\citep{valenciaLearningDistributionsComplex2024, yuLearningFlexibleBody2023} with shared layers.
We further introduce a time-parallel denoising scheme at inference, removing the speed bottleneck while maintaining accuracy.
Unlike diffusion-based \ac{cnn} simulators that require $K$ model evaluations per physical step~\citep{kohlBenchmarkingAutoregressiveConditional2024b, cachayDYffusionDynamicsinformedDiffusion, lippePDERefinerAchievingAccurate2023}, our method requires only a single hierarchical~\ac{gnn} call per step post warm-up, reducing inference costs over $T$ time steps from $\mathcal{O}(KT)$ to $\mathcal{O}(T)$. 
Compared to video-based approaches~\citep{ruheRollingDiffusionModels2024, gaoBayesianConditionalDiffusion2024}, which need to jointly process $N$ steps and learn a time-dependent denoiser with high memory cost, our model, \ac{robin}, leverages time-translation invariance to train a time-independent denoiser with only one time step in memory.
As such, \ac{robin} can be applied autoregressively and can freely interpolate between fully parallel denoising and memory-efficient one-step denoising at inference time.
It also predicts state residuals instead of states, significantly improving long-horizon rollout fidelity.

%% file: 00_sections/03_method.tex
\vspace{-0.1cm}
\section{~\acf{robin}} 
\label{sec:robin}
\vspace{-0.2cm}

\textbf{\acfp{gns} for Mesh-based Simulations.}
We consider solving~\acp{pde} for physical quantities $\mathbf{u}(\mathbf{x},t)$ that change over time $t \in [0,T]$ and inside a time-dependent domain $\mathbf{x}(t)\in \Omega(t)$. 
We focus on simulations on meshes, where $\mathcal{G} = (\mathcal{V},\mathcal{E}^{\text{M}})$ denotes the mesh graph and the graph nodes $\mathcal{V}$ and the graph edges $\mathcal{E}^{\text{M}}$ correspond to mesh nodes and mesh edges.
We seek solutions $\mathbf{u}_i(t) = \mathbf{u}(\mathbf{x}_i,t)$ at discrete node locations $\mathbf{x}_i(t)\in \Omega(t)$. 
To obtain discretized~\acp{pde}, usually numerical methods, such as the~\ac{fem}, are applied that define the discretization of spatial operators, such as gradients $\partial \mathbf{u}(\mathbf{x},t) / \partial \mathbf{x}$. 
Given the discretized operator $\mathcal{F}$, the~\ac{pde} simplifies to a time-dependent Ordinary Differential Equation and requires solving
$
    \partial \mathbf{u}_i /\partial t = \mathcal{F}(t,\mathbf{x}_i, \mathbf{u}_i)~\text{.}
$
We can solve such systems using numerical time discretization schemes. 
In this work we use a simple~\textit{Euler} forward discretization
$
    \mathbf{u}_i^{t{+}1} = \mathbf{u}_i^{t} + \Delta t~\mathcal{F}(t,\mathbf{x}_i^t, \mathbf{u}_i^t)~\text{,}
$
and set $\Delta t = 1$. 
We extend~\ac{pderef}~\citep{lippePDERefinerAchievingAccurate2023} to \textit{Lagrangian} systems, where the domain $\Omega^t$ and node locations $\mathbf{x}_i^t \in \Omega^t$ evolve over time.
Here, we predict the solution $\mathbf{u}_i^t$ at time step $t$ by learning to reverse a probabilistic diffusion process~\citep{hoDenoisingDiffusionProbabilistic2020a} conditioned on the previous state of time step $t-1$. 
The proposed methods also apply without any restriction to \textit{Eulerian} systems, where the domains are time-independent.


\subsection{\acfp{ddpm} for time-dependent simulations}
\label{subsec:ddpm}
\vspace{-0.1cm}

Given a time-dependent solution $\mathbf{u}^{t}$ from a data distribution $q(\mathbf{u})$, the forward diffusion process is modeled by a Markov chain, where Gaussian noise is added gradually to the sample $\mathbf{u}^{t}_k$ at each diffusion step $k$

\begin{equation*}
q(\mathbf{u}^t_{1:K}|\mathbf{u}^t_0) = \prod_{k=1}^K q(\mathbf{u}^t_{k}|\mathbf{u}^t_{k-1})~\text{,}~~~~q(\mathbf{u}^t_{k}|\mathbf{u}^t_{k-1}) := \mathcal{N}(\mathbf{u}^t_{k};\sqrt{1-\beta_k}\mathbf{u}^t_{k-1},\beta_k \mathbf{I})~\text{.}          
\end{equation*}

 $\mathcal{N}$ denotes the normal distribution and $\beta_k$ the noise specified by a variance scheduler. 
%
%
We learn to reverse the diffusion process, i.e., 

\begin{equation*}
p_{\theta}(\mathbf{u}^t_{k-1}|\mathbf{u}^t_{k}) := \mathcal{N}(\mathbf{u}^t_{k-1};\mu_{\theta}(\mathbf{u}^t_{k},k),\Sigma(\mathbf{u}^t_{k},k))~\text{,}    
\end{equation*}

consisting of $K$ iterative diffusion steps and starting from $k=K$. 
The mean $\mu_{\theta}$ depends on the prediction of a learned model with parameters $\theta$.
The covariance $\Sigma$ is assumed to be isotropic and given as $\Sigma = \sigma_k^2 \mathbf{I} = (\frac{1-\bar{\alpha}_{k-1}}{1-\bar{\alpha}_{k}}\beta_k) \mathbf{I}$~\citep{hoDenoisingDiffusionProbabilistic2020a} with $\alpha_i = 1 - \beta_i$ and $\bar{\alpha}_k = \prod_{i=1}^k \alpha_i$.
We train the model to predict the denoising velocity, i.e., the \textit{v-prediction target}

\begin{equation}
\label{eq:v_prediction}
\mathbf{v}_k^t = \sqrt{\bar{\alpha}_k} \mathbf{\epsilon}^t - \sqrt{1-\bar{\alpha}_k} \mathbf{u}^t_0~\text{,}
\end{equation}

given gaussian noise $\mathbf{\epsilon}^t$~\citep{salimansProgressiveDistillationFast2021}. 
Since this target smoothly interpolates between $-\mathbf{u}_0^t$ as $\bar{\alpha}_k \approx 0$ and $\boldsymbol{\epsilon}^t$ ($\bar{\alpha}_k \approx 1$), it emphasizes predicting the clean sample in early (high-noise) steps and the noise in later (high-signal) steps, simplifying learning.
The model predicts the denoising velocity $\mathbf{v}_\theta(\mathbf{u}^t_k, k) = \mathbf{v}_\theta(\mathbf{u}^t_k, k, \mathbf{u}^{t-1}_0)$ autoregressively, conditioned the model on the last time step solution $\mathbf{u}^{t-1}_0$.
The training objective is then defined as $\mathbb{E}_{\mathbf{u}^t_0, \mathbf{\epsilon}^t, k} \left[ \left\| \mathbf{v}_\theta(\mathbf{u}^t_k, k, \mathbf{u}^{t-1}_0) - \mathbf{v}_k^t \right\|^2 \right]$~\citep{lippePDERefinerAchievingAccurate2023}. 
To facilitate faster denoising, we follow the~\ac{ddpm} formulation of~\citep{lippePDERefinerAchievingAccurate2023} and use an exponential $\beta_k$ scheduler.

\input{05_inputs/figures/robin}

The first denoising step is defined such that $\bar{\alpha}_K\approx0$, which simplifies the v-prediction target to $\mathbf{v}_k^t \approx  - \mathbf{u}^t_0$ (cf. \Cref{eq:v_prediction}). The noisy sample $\mathbf{u}^{t}_{k} \approx \mathbf{\epsilon}^t$ corresponds to Gaussian Noise and is uninformative. 
Hence, for $k=K$ the model target converges to $\left\| \mathbf{v}_\theta(\mathbf{u}^t_K, K, \mathbf{u}^{t-1}_0)) - \mathbf{v}_k^t \right\|^2  \approx \left\| \mathbf{v}_\theta( \mathbf{\epsilon}^t, K, \mathbf{u}^{t-1}_0)) + \mathbf{u}^t_0\right\|^2$. 
This~\ac{mse} objective of the first denoising step $k=K$ mirrors the training objective of one-step models~\citep{lippePDERefinerAchievingAccurate2023}, i.e., auto-regressive models that predict the solution of the next time directly with a single prediction step.
Hence, we expect similar accuracy as one-step models during the first diffusion steps.  
However, the~\ac{ddpm}-based model refines the initial prediction iteratively at each time step, improving the accuracy of the solution further.
Note that, due to the noise scheduler, each denoising step focuses on different amplitude and frequency levels of the solution~\citep{lippePDERefinerAchievingAccurate2023}, with later denoising steps increasingly paying attention to higher frequencies.

\textbf{\acf{robi}.}
Conventional diffusion inference requires $K$ model calls, each corresponding to a denoising step, per simulation time step~\citep{lippePDERefinerAchievingAccurate2023}.
\Cref{fig:robin_methodology} a) shows an example.
Thus, inference is roughly $K$ times slower than one-step models~\citep{pfaffLearningMeshBasedSimulation2020}. 
We propose~\acf{robi} to accelerate inference in \ac{ddpm}-based autoregressive simulators.
Given a velocity prediction $\mathbf{v}_\theta(\mathbf{u}^{t}_{k}, k, \mathbf{u}^{t-1}_0)$ at the denoising step $k$, we reconstruct a partially refined prediction $
\tilde{\mathbf{u}}^{t}_{0|k} = \sqrt{\bar{\alpha}_k} \mathbf{u}^{t}_{k} - \sqrt{1-\bar{\alpha}_k} \mathbf{v}_\theta(\mathbf{u}^{t}_{k}, k, \mathbf{u}^{t-1}_0)
$, following~\citep{salimansProgressiveDistillationFast2021}.

As discussed in~\Cref{subsec:ddpm}, early denoising steps behave similarly to one-step model predictions, while later steps progressively refine higher spatial frequencies (from coarse structures to fine details).
After $m < K$ denoising steps at time $t$, the intermediate estimate $\tilde{\mathbf{u}}^{t}_{0|m}$ already captures low-to-mid frequency content that is sufficient to condition the next physical step $t{+}1$ to predict a solution within this already refined frequency band.

\ac{robi} exploits this property by starting the denoising of step $t{+}1$ as soon as step $t$ has progressed by $m$ steps. 
We initialize $\mathbf{u}^{t{+}1}_{K} \sim \mathcal{N}(\mathbf{0}, \mathbf{I})$ and denoise the batch of both time steps $t_w{\in}\{t,t+1\}$.
More generally, after $jm$ denoising steps, we initialize the time step $t{+}j$ and denoise a rolling time window with $t_w{\in}\{t,...,t+j\}$ in parallel. 
After a short warm-up, fully denoised samples (those with $k{=}0$) drop out and the window size becomes constant with $w{=}K/m$ and $t_w{\in}\{t,...,t{+}w{-}1\}$. 
Notably, the denoising index $k$ and thus the noise level increase along the window toward later physical times, aligning with the natural growth of forecast uncertainty. 
This reduces the number of model calls from $KT$ (conventional inference) to $K{-}m{+}mT$ steps.
For $m{=}1$, the number of model calls reduces to $K{+}T$, which is effectively the same complexity $\mathcal{O}(T)$ as for one-step models, when $K \ll T$.
We refer to $m$ as the~\textit{denoising stride}. \Cref{fig:robin_methodology} b) visualizes the special case $m{=}1$ and $K{=}3$ diffusion steps.
Each model call advances the simulation by one physical step and applies one denoising step to each of the $K$ partially denoised predictions in the window.

A single~\emph{Batched Denoising Step} (cf. \Cref{fig:robin_methodology} b) left for $m{=}1$ and $K{=}3$) of~\ac{robi} consists of two consecutive steps. 
In the~\emph{Prediction step}, the model outputs the velocity $\mathbf{v}_\theta(\mathbf{u}^{t_w}_{k_w}, k_w, \mathbf{u}^{t_w-1}_0)$ for each element in the prediction window $t_w{\in}\{t,...,t{+}w{-}1\}$, with the diffusion indices $k_w={\in}\{m,...,K\}$.
Subsequently, we perform a \emph{State Update} step.
Using the scheduler, we reconstruct $\tilde{\mathbf{u}}^{t_w}_{0|k_w}$ to update the conditioning states $\mathbf{u}^{t_w}_{0}$, and sample the reverse transition~$\mathbf{u}^{t_w}_{k_w-1} \sim p_{\theta}(\mathbf{u}^{t_w}_{k_w-1}|\mathbf{u}^{t_w}_{k_w})$. Both are used in the next~\textit{Prediction Step} as inputs. After $m$ such \emph{Batched Denoising Steps}, we advance in time and drop the fully denoised states ($k{=}0$) and initialize a new Gaussian sample ($k{=}K$) for the next time index~$t{+}w$, so the subsequent call evaluates $\mathbf{v}_\theta(\mathbf{u}^{t_w+1}_{k_w}, k_w, \mathbf{u}^{t_w}_{0})$. 

For $m{=}K$,~\ac{robi} reduces to conventional autoregressive diffusion inference (cf. \Cref{fig:robin_methodology} a)). 
Thus, the denoising stride $m$ can be considered a hyperparameter that trades off prediction accuracy and memory usage with inference speed.
Most notably, in \emph{State Update}, \ac{robi} reconstructs the physical states, while the rolling time window is treated purely as a batch dimension in the prediction model.
Consequently, the model always predicts~$\mathbf{v}_\theta(\mathbf{u}^t_k, k) = \mathbf{v}_\theta(\mathbf{u}^t_k, k, \mathbf{u}^{t-1}_0)$,
i.e., conditioned only on the previous reconstructed state during both training and inference, which proves to be more stable and accurate for autoregressive~\ac{ml}-based simulations~\citep{pfaffLearningMeshBasedSimulation2020,lippePDERefinerAchievingAccurate2023}.
Furthermore, this preserves the time-shift equivariance of Markovian systems, fast training convergence and small GPU memory utilization.

In practice, we find predicting residuals $\Delta\mathbf{u}^{t}_{k}$ improves accuracy.
Let the model denoise a batch of residual states $\Delta \tilde{\mathbf{u}}^{t_w}_{0} \approx \Delta \tilde{\mathbf{u}}^{t_w}_{0|k_w}$ of the time window $t_w{\in}\{t,...,t{+}w{-}1\}$ and denote the last fully denoised state as $\mathbf{u}^{t-1}_{0}$.
We then recover clean states inside the window via a cumulative sum $\tilde{\mathbf{u}}^{t+j}_{0} = \mathbf{u}^{t-1}_{0} + \sum_{i=0}^j \Delta \tilde{\mathbf{u}}^{t+i}_{0},~j\in\{0,...,w-1\}$. To further accelerate inference, we optionally stop denoising early at a \emph{truncation step} $k_\mathrm{tr}$ and use the partially denoised state $\tilde{\mathbf{u}}^{t}_{0|k_\text{tr}}$ as the final prediction of the time step. 
This remains effective because early denoising steps already approximate one-step prediction.

\subsection{\acfp{ddpm} for mesh-based simulations}
\vspace{-0.1cm}

To fully utilize \acfp{ddpm}'s potential for generating rich, multi-frequency solutions, prediction models must handle multi-scale information.
\acfp{hgnn} are particularly well-suited for this, as their architecture inherently learns representations at varying levels of granularity, mirroring the diverse frequency content present in~\ac{ddpm} outputs. Leveraging this idea, we train~\ac{hgnn} to predict the discrete denoising velocity $\mathbf{v}_{i,\theta}(\mathbf{u}^t_{i,k}, k, \mathbf{u}^{t-1}_{i,0})$ for mesh-based simulations on the mesh graph $\mathcal{G} = (\mathcal{V},\mathcal{E}^{\text{M}})$. It takes the current noisy sample $\mathbf{u}^{t}_{i,k}$ and is conditioned on the previous clean sample~$\mathbf{u}^{t-1}_{i,0}$.

\textbf{Root-node~\ac{amg}-based Mesh Coarsening.}
We construct a hierarchical mesh graph $\mathcal{G}^{\text{H}} = \mathcal{G}^{0:L} = (\mathcal{V}^{0:L},\mathcal{E}^{0:L,\text{M}})$ consisting of $L+1$ mesh graphs with nodes $\mathcal{V}^{0:L}$ and mesh edges $\mathcal{E}^{0:L,\text{M}}$ by coarsening the fine graph $\mathcal{G}^0:=\mathcal{G}$ $L$ times.
Coarsening is performed with root-node smoothed aggregation~\citep{vanekAlgebraicMultigridSmoothed1996b} implemented in~\citep{bellPyAMGAlgebraicMultigrid2022a}.
The solver takes the fine-mesh adjacency (with self-loops) $A^0$ as its system matrix and creates a hierarchy of adjacency $A^{0:L}$, upsampling (prolongation) $U^{0:L-1}$ and downsampling (restriction) $D^{0:L-1}$ matrices.
In practice, our implementation only requires the fine adjacency $A^0$, the root-node solver~\citep{bellPyAMGAlgebraicMultigrid2022a}, and constructing the graphs of each level from the non-zeros of the returned matrices $A^{1:L}$.
The resulting nodes $j \in \mathcal{V}^{0:L}$ of each level are always a subset of the fine mesh nodes.  
Unlike bi-stride pooling with Delaunay remeshing~\citep{yuLearningFlexibleBody2023}, this \ac{amg}-based approach better preserves mesh geometry by leveraging the strength of connections in the adjacency matrix.
\Cref{fig:bb_mesh_comparison} in \Cref{apx:results} visualizes this difference.

We additionally define $L-1$ downsampling edges $\mathcal{E}^{l,\text{D}}$ and upsampling edges $\mathcal{E}^{l,\text{U}}$, which connect nodes between successive levels $\mathcal{G}^{l}$ and $\mathcal{G}^{l+1}$.
We define these as the connections (non-zero values) of the given up- $U^{0:L-1}$ and downsampling $D^{0:L-1}$ matrices, resulting in an extended hierarchy graph $\mathcal{G}^{\text{H}} = (\mathcal{V}^{0:L},\mathcal{E}^{0:L,\text{M}} \cup \mathcal{E}^{0:L-1,\text{D}} \cup \mathcal{E}^{0:L-1,\text{U}})$.
As before, we found that the obtained pooling mappings respect the mesh geometry well, as shown in \Cref{fig:bb_mesh_comparison} d) in \Cref{apx:results}.

For contact experiments, we add contact edges $\mathcal{E}^{0,\text{C}}$~\citep{yuLearningFlexibleBody2023} to the graph hierarchy $\mathcal{G}^{\text{H}}$.
In a simulation involving two colliding components, we define a bidirectional edge between the nodes of the first part and the nodes of the second part if their distance is less than the specified contact radius $R$.
The resulting hierarchy is given by \mbox{$\mathcal{G}^{\text{H}} = (\mathcal{V}^{0:L},\mathcal{E}^{0:L,\text{M}}\cup \mathcal{E}^{0:L,\text{C}}\cup \mathcal{E}^{0:L-1,\text{D}}\cup \mathcal{E}^{0:L-1,\text{U}})$}.

\subsection{\acfp{ampn}.}

\textbf{Encoder.} 
Let $\mathcal{G}^{\text{H}}$ be a hierarchical graph as defined above and $\mathbf{u}^{t}_{i,k}$ the noisy sample at denoising step $k$ of simulation step $t$.  
We define node embeddings $\mathbf{k}_i \in \mathcal{V}^{0:L}$, mesh edge embedding $\mathbf{e}_{ij}^{\text{M}} \in \mathcal{E}^{0:L,\text{M}}$, contact edge embeddings $\mathbf{e}_{ij}^{\text{C}} \in \mathcal{E}^{\text{C}}$, downsampling edge embeddings $\mathbf{e}^{\text{D}}_{ij} \in \mathcal{E}^{0:L-1,\text{D}}$ and upsampling edge embeddings $\mathbf{e}^{\text{U}}_{ij} \in \mathcal{E}^{0:L-1,\text{U}}$.
We add relative node distances $\mathbf{x}^t_{ij} = \mathbf{x}_i^t-\mathbf{x}_j^t$ and their norm $|\mathbf{x}^t_{ij}|$ to all edge embeddings and the initial distance $\mathbf{x}^0_{ij} = \mathbf{x}^0_{i} - \mathbf{x}^0_{j}$ and their norm $|\mathbf{x}^0_{ij}|$ to mesh edges, down- and upsampling embeddings. 
Node embeddings include a one-hot encoding $n_i$ of the node type.
Node embeddings at level $l=0$ additionally include $\mathbf{u}^{t}_{i,k}$ and task-specific features.  
All embeddings are projected to the latent dimension $d$ via linear layers.  
We add a Fourier encoding~\citep{hoDenoisingDiffusionProbabilistic2020a} for the denoising step $k$ and the normalized level $l^* = l/L$ to inform the~\ac{ampn} of relative graph depth.

\textbf{Processor and Decoder.} Similar to~\ac{amg} solvers~\citep{ruge4AlgebraicMultigrid1987,vanekAlgebraicMultigridSmoothed1996b} and  UNets~\citep{ronnebergerUNetConvolutionalNetworks2015},~\acfp{ampn} use a~\textit{V-cycle} to propagate information between levels.
They consist of five core message passing modules: Pre-Processing, Downsampling, Solving, Upsampling and Post-Processing, as shown in \Cref{fig:hierarchical_graph_diffusion}.
Pre-Processing, Solving and Post-Processing modules use an~\ac{intramps} consisting of $N$ message passing steps to update the heterogeneous subgraph $\mathcal{G}^l = (\mathcal{V}^l,\mathcal{E}^{l,\text{C}}\cup \mathcal{E}^{l,\text{M}})$ of level $l$. 
Given node embeddings $\mathbf{k}_i \in \mathcal{V}^l$, contact edge embeddings $\mathbf{e}_{ij}^\text{C} \in \mathcal{E}^{l,\text{C}}$ and mesh edge embeddings $\mathbf{e}_{ij}^{\text{M}} \in \mathcal{E}^{l,\text{M}}$, the message passing update of the level graph at step $n$ is defined by

\begin{equation} \label{eqn:mpn}
\begin{aligned}
\mathbf{e}_{ij}^{\text{C},n+1} &= W^{n}_{\theta,\mathcal{E}^{\text{C}}} ~\mathbf{e}_{ij}^{\text{C},n} + f^{n}_{\theta,\mathcal{E}^{\text{C}}}(\mathbf{k}^n_i, \mathbf{k}^{n}_j, \mathbf{e}_{ij}^{\text{C},n}) \textrm{ , } \\
\mathbf{e}_{ij}^{\text{M},n+1} &=  W^{n}_{\theta,\mathcal{E}^{\text{M}}} ~\mathbf{e}_{ij}^{\text{C},n} + f^{n}_{\theta,\mathcal{E}^{\text{M}}}(\mathbf{k}^{n}_i, \mathbf{k}^{n}_j, \mathbf{e}_{ij}^{\text{M},n}) \textrm{ , }\\
\mathbf{k}^{n+1}_i &= W^{n}_{\theta,\mathcal{V}} ~\mathbf{k}^{n}_i + f^{n}_{\theta,\mathcal{V}}(\mathbf{k}^{n}_i,  \bigoplus_{j} \mathbf{e}_{ij}^{\text{C},n+1} ,  \bigoplus_{j} \mathbf{e}_{ij}^{\text{M},n+1} )\quad \text{.} \\
\end{aligned}
\end{equation}

The operator $\bigoplus$ denotes a permutation-invariant aggregation, $f^{n}_{\theta,.}$~\acp{mlp} and $W^{n}_{\theta,.}$ weight matrices~\citep{battagliaRelationalInductiveBiases2018a,pfaffLearningMeshBasedSimulation2020, valenciaLearningDistributionsComplex2024}.
Downsampling modules update the subgraph $\mathcal{G}^{l,\text{D}} = (\mathcal{V}^{l+1}\cup \mathcal{V}^l, \mathcal{E}^{l,\text{D}})$ with an~\ac{intermps} of $N$ message passing steps.
The receiver embeddings are $\mathbf{k}^{\text{rec}}_i \in \mathcal{V}^{l+1}$, the sender embeddings $\mathbf{k}^{\text{send}}_j \in \mathcal{V}^l$, and the edge embeddings $\mathbf{e}_{ij} \in \mathcal{E}^{l,\text{D}}$.
Similarly, the upsampling layers update the embeddings of the subgraph $\mathcal{G}^{\text{U}}_{l} = (\mathcal{V}^{l+1}\cup  \mathcal{V}^l, \mathcal{E}^{l,\text{U}})$ of level $l$.
Here, the receiver embeddings are $\mathbf{k}^{\text{rec}}_i \in \mathcal{V}^{l}$, the sender embeddings $\mathbf{k}^{\text{send}}_j \in \mathcal{V}^{l+1}$, and the edge embeddings $\mathbf{e}_{ij}  \in \mathcal{E}^{l,\text{U}}$.
A message passing step of an~\ac{intramps} is defined as
  
\begin{equation} \label{eqn:mpn}
\begin{aligned}
\mathbf{e}_{ij}^{n+1} &= W^{n}_{\theta,\mathcal{E}} ~\mathbf{e}_{ij}^{n} + f^{n}_{\theta,\mathcal{E}}(\mathbf{k}^{\text{rec},n}_i, \mathbf{k}^{\text{send},n}_j, \mathbf{e}_{ij}^{n}) \textrm{ , } \\
\mathbf{k}^{\text{rec},n+1}_i &= W^{n}_{\theta,\mathcal{V}} ~\mathbf{k}^{\text{rec},n}_i + f^{n}_{\theta,\mathcal{V}}(\mathbf{k}^{\text{rec},n}_i,  \bigoplus_{j} \mathbf{e}_{ij}^{n+1})\quad \text{.} \\
\end{aligned}
\end{equation} 
\vspace{-0.3cm}

Our~\textit{V-cycle} starts at level $l=0$ with pre-processing and downsampling at each level, repeated until the coarsest level $l=L$ is reached. 
We then apply multiple message passing steps at level $L$, which has the largest receptive field.  
Next, we upsample and post-process each level back up to $l=0$.  
All Pre-Processing, Downsampling, Upsampling, and Post-Processing modules share weights across levels.  
A final linear layer decodes the fine-level node embeddings $\mathbf{k}_i \in \mathcal{V}^0$ to produce the velocity prediction $\mathbf{v}_{i,\theta}(\mathbf{u}^t_{i,k}, k, \mathbf{u}^{t-1}_{i,0})$.

%% file: 05_inputs/figures/robin.tex
\begin{figure}[!htb]
    \centering
    \includegraphics[width=\textwidth,trim={20pt 150pt 0pt 190pt},  clip,]{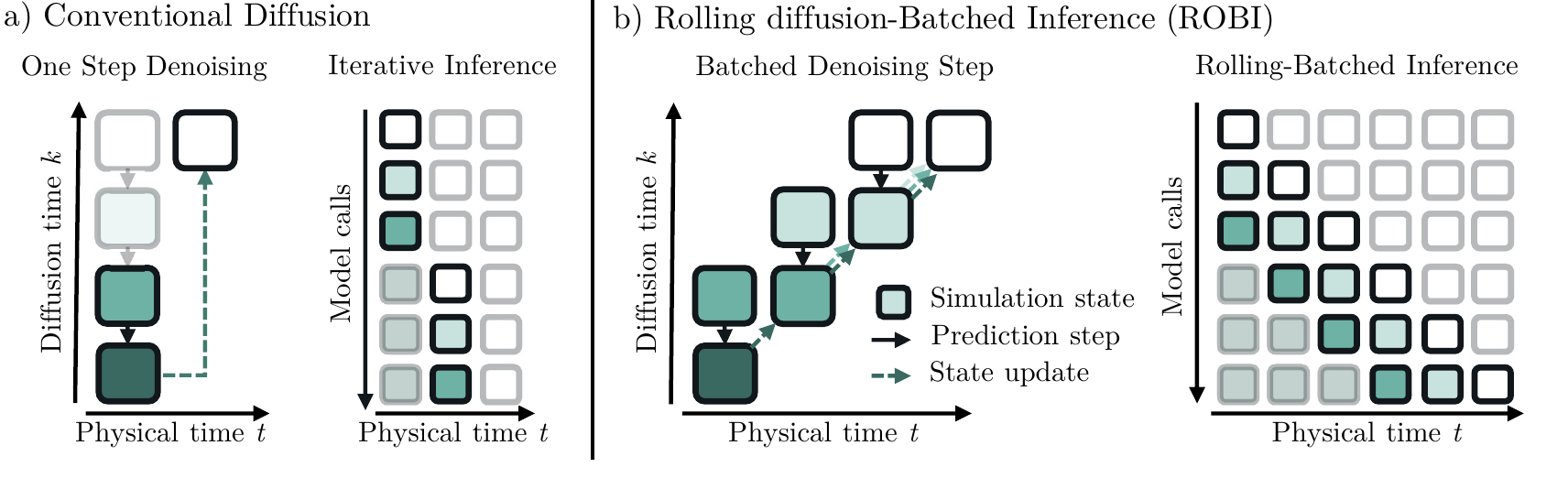}
    \caption{Overview of \textbf{a)} conventional autoregressive diffusion inference and \textbf{b)} \ac{robi}. 
    \textbf{a)} Conventional inference denoises the entire state of a physical time step at once before it shifts to the next time step (see \textit{One Step Denoising}). 
    The~\textit{Iterative Inference} requires $K$ model calls per time step, where $K$ denotes the number of diffusion steps.
    \textbf{b)} \ac{robi} parallelizes denoising steps across physical time, processing up to $K$ time steps batched, and reconstruct the physical states with the clean sample prediction subsequently (see \textit{Batched Denoising Step}). 
    This process allows \textit{Rolling-Batched Inference} after the initial warm-up, reducing the number of model calls to one per time step.
    }
    \label{fig:robin_methodology}
\end{figure}

%% file: 00_sections/04_experiments.tex
\vspace{-0.1cm}
\section{Experiments}
\label{sec:experiments}
\vspace{-0.2cm}

\textbf{Datasets.}
We evaluate our model on the three different datasets, namely \textsc{BendingBeam}, \textsc{ImpactPlate}~\citep{yuLearningFlexibleBody2023} and \textsc{DeformingPlate}~\citep{pfaffLearningMeshBasedSimulation2020}.
We introduce the \textsc{BendingBeam} dataset (\Cref{fig:overview_experiments}a)), featuring quasi-static, geometrically non-linear deformations of beams with high aspect ratios. 
The setup challenges models to capture global deformations via broad receptive fields and resolve high spatial frequencies due to locally thin, low-stiffness regions.
In \textsc{ImpactPlate} (\Cref{fig:overview_experiments} b)), the models must learn flexible dynamics with varying material parameters and accurately resolve very localized deformation at the contact point.
\textsc{DeformingPlate} (\Cref{fig:overview_experiments} c)) considers quasi-static contact simulations induced by scripted actuators that deform $3$D plates consisting of nonlinear, hyperelastic material.

\input{05_inputs/figures/overview_experiments}


\textbf{Experimental setup.}
For all tasks, we target the displacement residual of the node positions with respect to the next time step $\Delta \mathbf{u}^t_{i,0} = \mathbf{x}^{t+1}_{i} - \mathbf{x}^{t}_{i}$ during denoising. 
We additionally denoise the von Mises stress $\sigma^t_{\text{vMises},i,0}$ directly without residuals to gain further insight into the dynamics of the three experiments.
\ac{robin} uses $K=20$ denoising steps and a denoising stride of $m=1$ by default. 
The task-specific features are specified in \Cref{apx:setup}, listed in \Cref{table:features}.
We measure the prediction error using an \textit{RMSE}, as specified in~\Cref{apx:setup}.
\Cref{apx:setup} also provides information about additional settings of~\ac{robin}, including training details and hyperparameters.

\textbf{Baselines.} 
We compare our model with three prominent baselines for nonlinear deformation simulations, namely~\acp{mgn},~\acp{hcmt}, and~\acp{bsms}. 
Detailed setups are provided in~\Cref{apx:baselines}. 

\textbf{Variants.}
We demonstrate that a \emph{single trained}~\ac{robin} can easily switch between different rollout modes by varying the denoising stride $m$ and the truncation step $k_\text{tr}$. 
We therefore evaluate multiple rollout settings ($m$/$k_\text{tr}$) on the same trained model~\ac{robin} model: the default (1/20), conventional autoregressive diffusion inference (20/20) (i.e., $m{=}20$ denoising steps per physical step), and an intermediate variant (5/20) with $m=5$.
To study early stopping, we further reduce the truncation step $k_\text{tr}$ with (1/10), (1/5), (1/3), (1/2) and (1/1).

\textbf{Ablations.}
We ablate key components of~\ac{robin} to assess their impact.  
\emph{10 diffusion steps} and~\emph{5 diffusion steps} train with reduced diffusion length $K$.
\emph{W/o diffusion} trains the~\ac{ampn} as a one-step autoregressive model with~\ac{mse} loss.  
\emph{W/o hierarchy} disables hierarchical message passing, operating only on the fine mesh $G^0$.  
\emph{State prediction} replaces residual-based prediction with direct denoising of $\mathbf{u}^t_{i,k}$ instead of $\Delta \mathbf{u}^t_{i,k}$.  
\emph{W/o shared layer} uses $15$ unshared message passing layers.
\emph{HCMT model} replaces~\acp{ampn} by~\acp{hcmt} as the hierarchical model for~\ac{robin}.
\Cref{apx:baselines} provides additional implementation and training details.

\textbf{Generalization to large meshes.} 
To assess upscaling, we introduce \textsc{BendingBeamLarge} with meshes on average over ten times larger than in \textsc{BendingBeam}.
We compare fine-tuning a pre-trained \ac{robin} to training from scratch, showing rapid transfer to substantially larger meshes without architectural changes, enabled by shared blocks in the \acp{ampn} that accommodate varying hierarchy depths.

%% file: 05_inputs/figures/overview_experiments.tex
\begin{figure}[!htb]
    \centering
    \includegraphics[width=0.8\textwidth,trim={150pt, 110pt, 100pt, 200pt},  clip,]{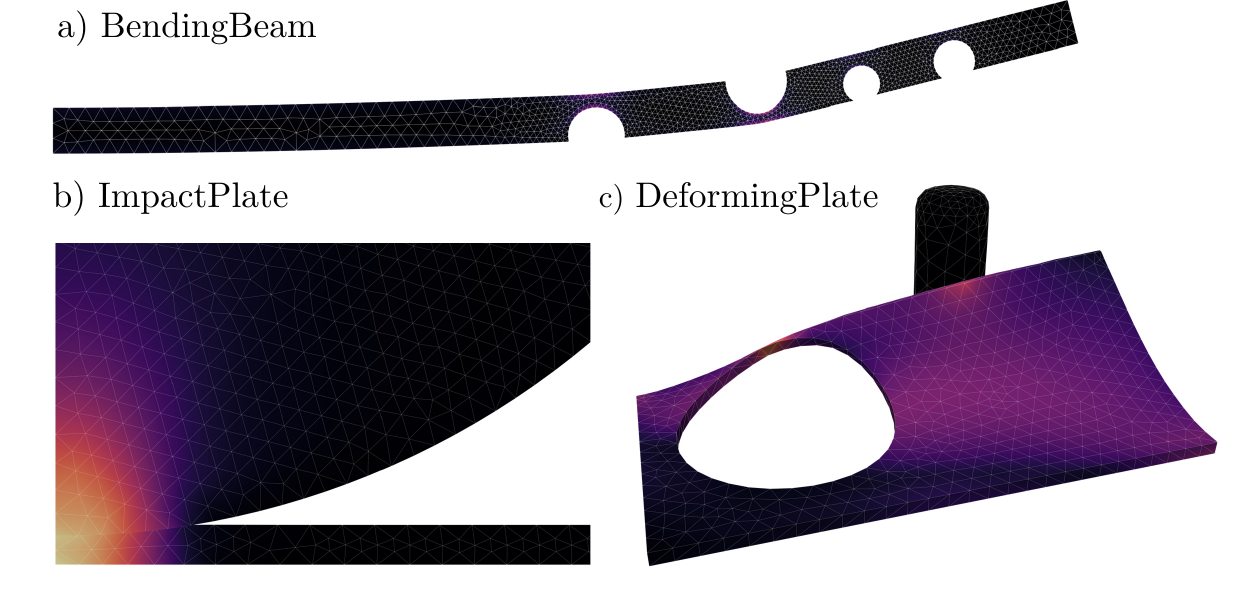}
    \vspace{-0.3cm}
    \caption{Example predictions of~\ac{robin} on the considered datasets.~\ac{robin} predicts the part deformations as well as the von Mises stress (color, yellow is high) on all experiments. 
    \textbf{a)} \textsc{BendingBeam} considers global part deformations of beams induced by local forces.
    \textbf{b)} In~\textsc{ImpactPlate}, the models have to predict locally large deformations, caused by a collision with an accelerated ball.
    \textbf{c)} The hyperelastic plates in~\textsc{DeformingPlate} are deformed by a scripted actuator. 
    }
    \label{fig:overview_experiments}
    \vspace{-0.3cm}
\end{figure}

%% file: 00_sections/05_results.tex
\vspace{-0.1cm}
\section{Results}
\label{sec:results}
\vspace{-0.2cm}

\textbf{Baselines.}
\Cref{fig:comb_baseline_comparison_pareto} a) compares the rollout RMSE of \ac{robin} against~\ac{hcmt}, ~\ac{mgn}, and~\ac{bsms}. 
\ac{robin} yields substantial error reductions for the prominent \textsc{ImpactPlate} and \textsc{DeformingPlate} datasets, and yields an even larger improvement on \textsc{BendingBeam}. \Cref{fig:res_qual_bb_comparison} demonstrates that~\ac{robin} is able to propagate local boundary conditions across the part for accurate predictions of the global part deformation.
In addition,~\ac{robin} resolves the non-linear bending curve of the~\ac{fem}. 
This requires fusing geometric features across scales, e.g., global part dimensions together with local thin walls, for an accurate prediction of the global part stiffness. 
All baselines struggle to reproduce the~\ac{fem} results, particularly the global deformation modes. 

\input{05_inputs/figures/combined_baseline_and_inference}
\input{05_inputs/figures/bb_comparison}

\textbf{Inference speed.}
\Cref{fig:comb_baseline_comparison_pareto} b) plots rollout RMSE versus inference time on~\textsc{BendingBeam}.
The default variant (1/20) of \ac{robin} is about an order of magnitude faster than conventional autoregressive diffusion inference (20/20), without compromising accuracy. 
Decreasing the truncation step $k_\text{tr}$ (i.e., fewer diffusion steps per physical step at inference) trades rollout accuracy for speed.
Nevertheless, the fastest variant (1/1), which can be seen as a one-step model variant of~\ac{robin}, still significantly outperforms the accuracy of the baselines (cf. \Cref{fig:comb_baseline_comparison_pareto} a)).
Most notably,~\ac{robin} (1/20) requires only $\approx 58\%$ more time than the one-step variant (1/1), despite performing $K{=}20$ denoising steps per time step, highlighting the efficiency  of the parallel denoising scheme~\ac{robi}.
Similar trends hold for~\textsc{ImpactPlate} and~\textsc{DeformingPlate} (cf.~\Cref{fig:ap_pareto} in~\Cref{apx:results}).
\Cref{fig:res_truncation_qual} in~\Cref{apx:results} visualizes how high frequency errors accumulate over the rollout if we skip the later denosing steps, demonstrating the importance of reducing the high frequency errors for long rollouts despite low displacement RMSE. 

\input{05_inputs/figures/truncation}
\textbf{Diffusion truncation.}~\Cref{fig:res_truncation} visualizes displacement RMSE and fine-mesh edgewise gradient RMSE $||(\mathbf{u}_{i} - \mathbf{u}_{j})||_2 / ||(\mathbf{x}^0_{i} - \mathbf{x}^0_{j})||_2$, against ground truth. 
Early diffusion steps primarily remove low-frequency error (global RMSE drops), whereas later steps reduce high-frequency error (gradient RMSE).
Skipping the later denoising steps results in high frequency error accumulation and mesh degradation throughout the rollout (cf. \Cref{fig:res_truncation_qual} in \Cref{apx:results}), as observed in one step models~\citep{lippePDERefinerAchievingAccurate2023}.
With $m{=}1$, \ac{robin} does not increase the gradient error on~\textsc{BendingBeam}.
We observe similar results on \textsc{DeformingPlate} and \textsc{ImpactPlate}, as illustrated in \Cref{fig:res_truncation_ip_dp} in~\Cref{apx:results}.


\textbf{Ablations.} As shown in \Cref{fig:res_ablations}, ablations confirm the importance of each \ac{robin} component. 
Reducing the diffusion length $K$ slightly decreases accuracy across all datasets. 
However, even with $K{=}5$ diffusion steps,~\ac{robin} remains significantly more accurate than all baselines.
Using non-shared layers significantly degrades performance on \textsc{BendingBeam}, likely due to a reduced effective receptive field.
The large error increase for \emph{State prediction} indicates that residual prediction is crucial for~\textsc{Lagrangian} simulations.
Across all datasets,~\ac{robin} outperforms the non-hierarchical variant, the non-diffusion variant, and the combination of~\acp{ddpm} with the strongest hierarchical baseline (\ac{hcmt}), demonstrating the synergy between~\acp{ddpm} and~\acp{ampn}.

\input{05_inputs/figures/ablations}

\textbf{Generalization to large meshes.} Fine-tuning the pre-trained~\ac{robin} converges markedly faster and attains substantially lower RMSE than training from scratch, demonstrating mesh-size independence and efficient transfer to deeper~\ac{amg} hierarchies with twelve times more nodes, without architectural changes. 
In~\Cref{fig:res_generalization}, pretrained predictions are visually indistinguishable from~\ac{fem} even in thin, high-stress regions with large bending.
The model trained from scratch exhibits negligible deformation due to much slower convergence within the same training budget.
~ \ac{robin} consistently requires $31$~s per case, while the numerical solver averages~$108$~s and can require up to $4248$~s, due to problem-dependent nonlinear convergence costs. 
\Cref{table:results_table} in~\Cref{apx:results} lists the quantitative results. 

\input{05_inputs/figures/generalization}

%% file: 05_inputs/figures/combined_baseline_and_inference.tex
\begin{figure}[htbp]
    \centering 
    \vspace{-0.25cm}

    \begin{subfigure}[b]{0.45\textwidth} 
        \centering 
        \includegraphics[width=\textwidth,trim={5pt, 10pt, 5pt, 10pt},  clip]{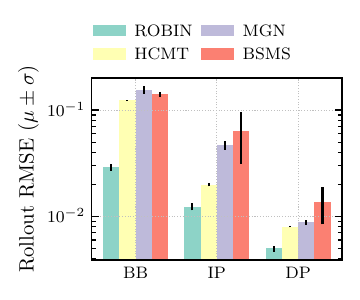}
        \label{fig:baseline_comparison}
    \end{subfigure}
    \hfill 
    \begin{subfigure}[b]{0.53\textwidth} 
        \centering 
        \includegraphics[width=\textwidth,trim={5pt, 5pt, 5pt, 12pt},  clip]{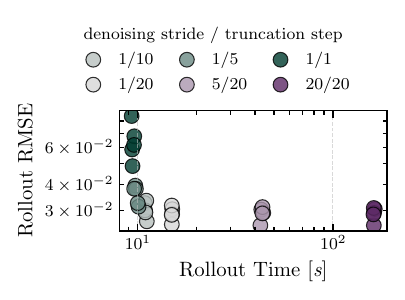}
        \label{fig:pareto}
    \end{subfigure}
    \vspace{-0.3cm}
    \caption{\textbf{Left:} Rollout error measured by the RMSE of the predicted nodes positions.~\ac{robin} surpasses the accuracy of the baselines~\ac{hcmt},~\ac{mgn}, and~\ac{bsms} on all three datasets~\textsc{BendingBeam},~\textsc{ImpactPlate} and ~\textsc{DeformingPlate}. 
    \textbf{Right:} 
    Comparison of inference time and error of~\ac{robin} and its variants on \textsc{BendingBeam}.
    The default variant of \ac{robin} (1/20) achieves the same accuracy as conventional diffusion inference (20/20), while the inference speed is close to the one step variant (1/1). Reducing the truncation step $k_\text{tr}$ trades accuracy for speed.}
    \label{fig:comb_baseline_comparison_pareto}
    \vspace{-0.3cm}
\end{figure}


%% file: 05_inputs/figures/bb_comparison.tex
\begin{figure}[!htb]
    \centering
    \includegraphics[width=1.0\textwidth,trim={60pt, 140pt, 60pt 150pt},  ]{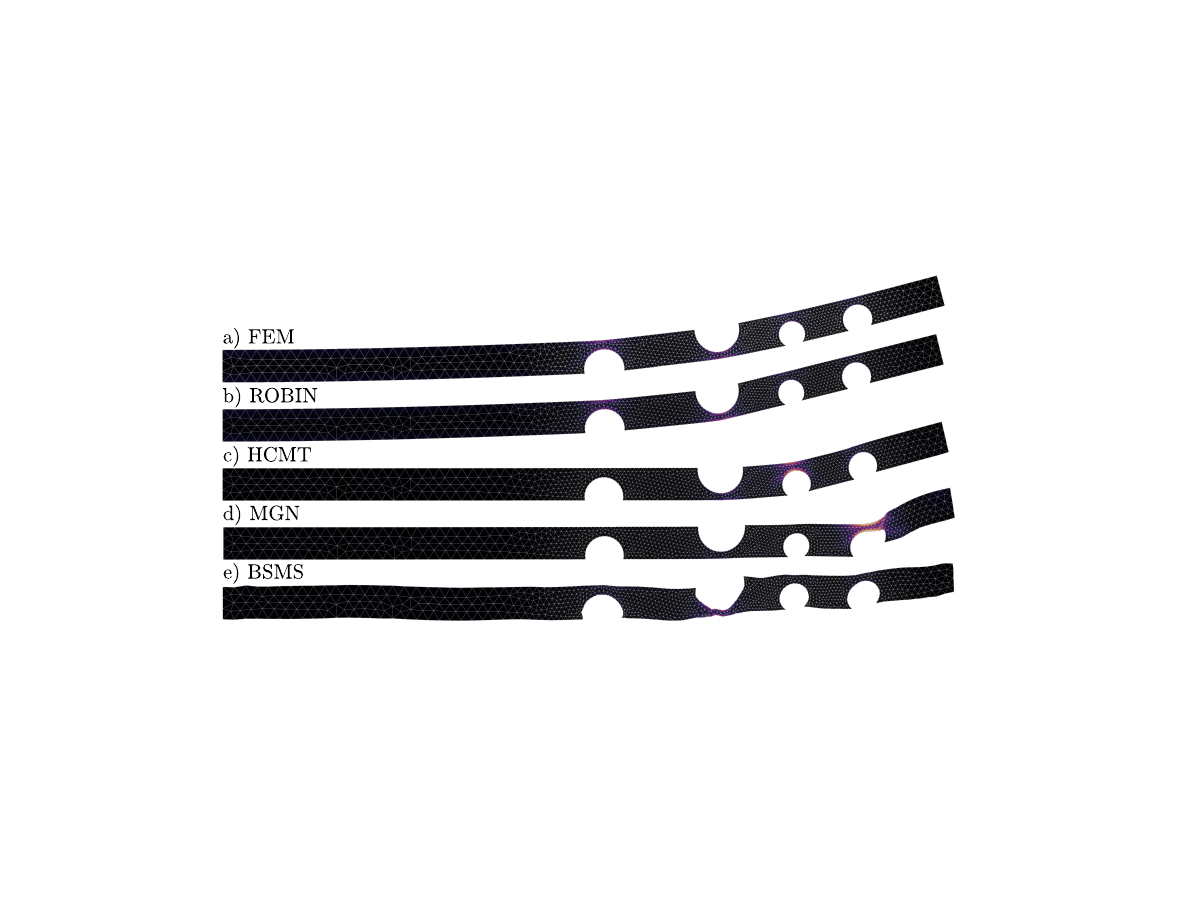}
    \caption{
    Comparison of the predicted rollout deformations and von Mises stresses (color, yellow is high) on \textsc{BendingBeam} between 
    \textbf{a)} the \ac{fem},
    \textbf{b)} \ac{robin}, 
    \textbf{c)} \ac{hcmt}, 
    \textbf{d)} \ac{mgn}, and  
    \textbf{e)} \ac{bsms}.  
    \ac{robin} is able to accurately reproduce the~\ac{fem} results. 
    Neither~\ac{hcmt},~\ac{mgn} nor~\ac{bsms} are able to resolve global deformation modes, illustrating the importance of the~\ac{ampn} for global message propagation.
    }
    \label{fig:res_qual_bb_comparison}
    \vspace{-0.6cm}
\end{figure}

%% file: 05_inputs/figures/truncation.tex

\begin{wrapfigure}{hR}{0.55\textwidth}
    \vspace{-0.0cm}
    \centering
    \includegraphics[width=0.5\textwidth,trim={8pt, 5pt, 0pt 10pt}, clip ]{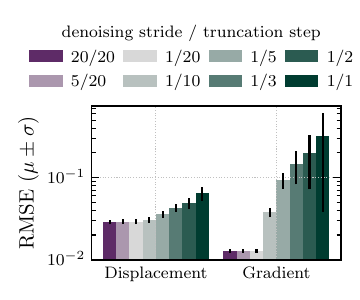}
    \caption{
    Comparison of the predicted rollout displacements and displacement gradients on \textsc{BendingBeam} for different truncation steps $k_\text{tr}$. 
    While the first diffusion steps strongly decrease the displacement RMSE, the later steps are important to reduce the local displacement gradient RMSE. 
    A fast inference with a low denoising stride neither increases the displacement RMSE nor the gradient RMSE during a rollout. 
    }
    \label{fig:res_truncation}
    \vspace{-0.5cm}
\end{wrapfigure}

%% file: 05_inputs/figures/ablations.tex
\begin{figure}[!htb]
    \centering
    \includegraphics[width=\textwidth,trim={5pt, 5pt, 5pt, 10pt},  clip,]{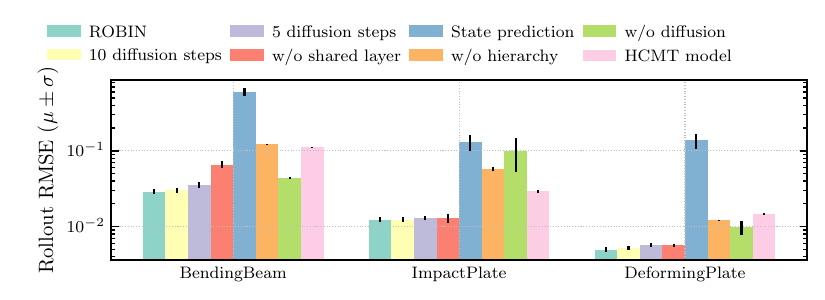}
    \caption{Rollout RMSE of the displacement predictions.~\ac{robin} remains accurate, even for a very small number of diffusion steps. Shared layers are crucial for~\textsc{BendingBeam} to increase the receptive field.
    Replacing residual predictions with state predictions, hierarchical architectures with non-hierarchical architectures or with~\ac{hcmt} architectures, and diffusion with non-diffusion architectures significantly decrease the accuracy across all datasets. }
    \label{fig:res_ablations}
    \vspace{-0.3cm}
\end{figure}

%% file: 05_inputs/figures/generalization.tex
\begin{figure}[!htb]
    \centering
    \includegraphics[width=\textwidth,trim={50pt, 30pt, 40pt 8pt},clip  ]{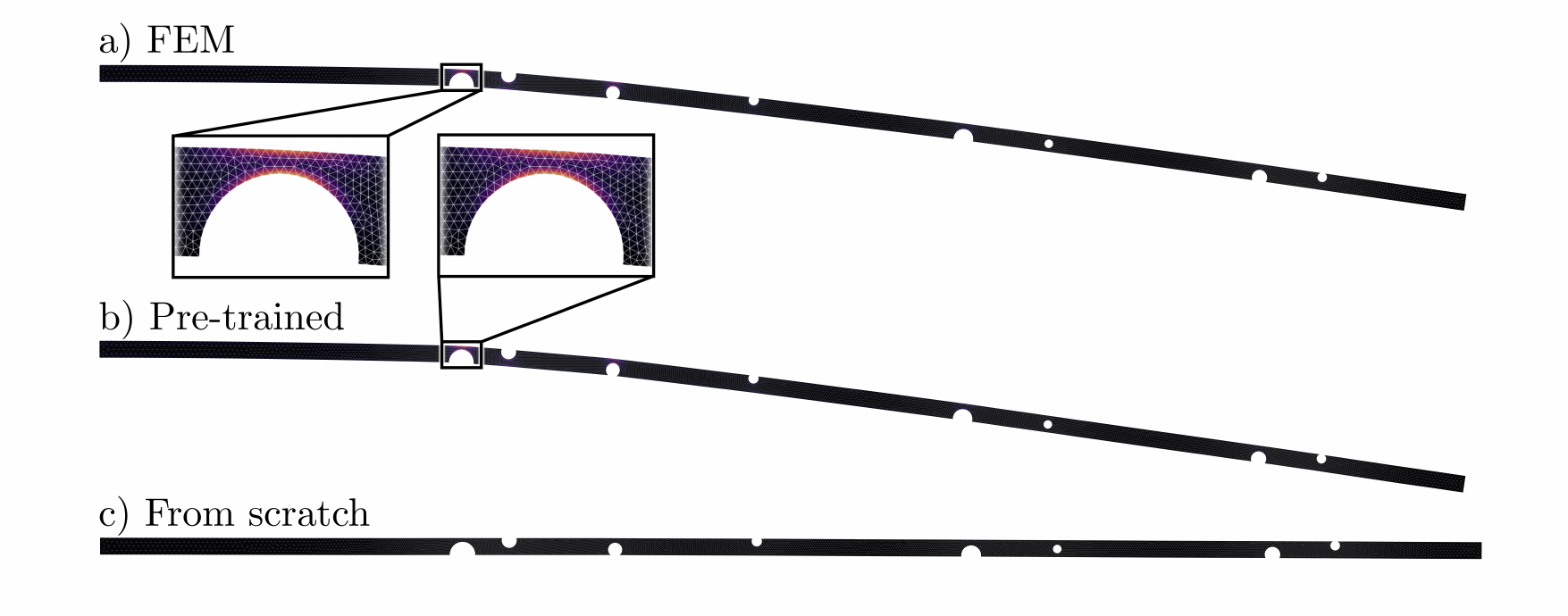}
    \caption{
    Comparison of the predicted rollout deformations and von Mises stresses (color, yellow is large) on~\textsc{BendingBeamLarge} between 
    \textbf{a)} the \ac{fem},
    \textbf{b)} a fine-tuned~\ac{robin} model, which was pre-trained on the small~\textsc{BendingBeam} dataset, and, 
    \textbf{c)} a~\ac{robin} model trained from scratch for the same number of training iterations.
    The fine-tuned~\ac{robin} closely matches the~\ac{fem} deformation and stress distribution, capturing even local high-stress hotspots, whereas the from-scratch model underestimates deformation due to much slower convergence within the same training budget.
    }
    \label{fig:res_generalization}
    \vspace{-0.3cm}
\end{figure}

%% file: 00_sections/06_conclusion.tex
\vspace{-0.1cm}
\section{Conclusion}
\vspace{-0.2cm}

We introduced \acf{robin}, a diffusion-based \ac{hgnn} that utilizes \acp{ampn} to refine mesh-based predictions across scales. 
Leveraging the expressiveness of multiscale message passing and the accuracy of diffusion, \ac{robin} outperforms state-of-the-art simulators on varied nonlinear solid mechanics tasks in terms of predictive accuracy.
These tasks include a novel~\textsc{BendingBeam} dataset that reveals limitations of current learned simulators.
\ac{robi}, \ac{robin}'s inference scheme, parallelizes diffusion across time steps, reducing inference runtime by up to an order of magnitude without sacrificing accuracy. 
We validated \ac{robin} on three challenging datasets, including the new \textsc{BendingBeam} benchmark, and demonstrated significant gains in accuracy and efficiency.
We discuss the broader impact of this work and our method in Appendix~\ref{apx:broader_impact}.

\textbf{Limitations and Future Work.} 
\ac{robin} currently does not possess SO(3) equivariance. Adding this property could improve the accuracy of orientation-sensitive predictions. 
We focus on \ac{ddpm}, while other diffusion formulations or denoising schedules could provide valuable insights and further enhance performance.
Similarly, our experiments cover nonlinear solid mechanics, and extensions to other domains, such as fluid dynamics, are a promising direction. Lastly, \ac{robin}'s combination of fast inference and high accuracy opens opportunities for accelerating multi-stage design and optimization workflows.

%% file: 00_sections/07_acknowledgements.tex
\section*{Acknowledgements}
This work is part of the DFG AI Research Unit 5339 regarding the combination of physics-based simulation with AI-based methodologies for the fast maturation of manufacturing processes. The financial support by German Research Foundation (DFG, Deutsche Forschungsgemeinschaft) is gratefully acknowledged. The authors acknowledge support by the state of Baden-Württemberg through bwHPC, as well as the HoreKa supercomputer funded by the Ministry of Science, Research and the Arts Baden-Württemberg and by the German Federal Ministry of Education and Research. This work is supported by the Helmholtz Association Initiative and Networking Fund on the HAICORE@KIT partition.

%% file: 02_appendix/0X_broader_impact.tex
\section{Broader Impact}
\label{apx:broader_impact}

The~\ac{ml}-based simulator,~\acf{robin}, offers significant advantages for computational modeling and simulation. This is achieved by reducing computational costs while maintaining accuracy. This enables engineers to iterate through significantly more design variations or to quickly evaluate numerous scenarios using the fast model. However, like all powerful computational tools, there is a risk of misuse, for instance, in weapons development or unsustainable resource exploitation.

%% file: 02_appendix/0X_datasets.tex
\section{Datasets}\label{apx:datasets}

\Cref{table:datasets} provides an overview of the considered datasets in this work.

\input{05_inputs/tables/datasets}

\textbf{{\textsc{BendingBeam}}.} This dataset considers the bending of beam parts due to external forces. 
Bending is one of the most basic deformation modes of parts in structural mechanics. 
The dataset is designed as a diagnostic benchmark for neural~\ac{pde} solvers, addressing various potential bottlenecks. 
The force and handle boundary conditions are very local, only being defined on a small subset of mesh nodes (cf. \Cref{fig:bb_force_bc}).
However, the resulting deformations affect all nodes.

\input{05_inputs/figures/bb_force_bc}

Hence, the neural~\ac{pde} must effectively propagate local information to all nodes of the mesh. 
Next, the dataset considers beams with large aspect ratios. 
This results in large graph diameters, which represent the shortest path between the most distant nodes. 
The mesh resolution is increased at thin walls, which additionally increases the graph diameter. 
The local boundary conditions have to be transmitted across a large number of nodes, which challenges the ability to propagate messages globally.
The geometry and especially the thin locations of the geometry strongly influence the global bending stiffness and deformation of the part.
Overall, the model has to output accurate solutions across various spatial frequencies. 

The solutions are created with scikit-fem~\citep{gustafsson2020scikit}, iteratively solved using Newton-Raphson until the residual fell below a tolerance of $10^{-8}$.
Each simulation is solved for a total number of $400$ time steps.
We create a total number of $1000$ simulations for training, $100$ for validation and $100$ for testing. 

To evaluate the generalization capability of~\ac{robin}, we create an additional dataset variant~\textsc{BendingBeamLarge}, containing meshes more than ten times larger than those in ~\textsc{BendingBeam}, with an average of $8897$ nodes. One simulation in~\textsc{BendingBeamLarge} contains $100$ time steps. 
    

%% file: 05_inputs/tables/datasets.tex
\newcolumntype{k}{>{\centering\arraybackslash\hsize=.2\hsize}X}
\newcolumntype{b}{>{\centering\arraybackslash\hsize=.3\hsize}X}
\begin{table}[H]
\caption{Comparison of the datasets~\textsc{BendingBeam},~\textsc{BendingBeamLarge},~\textsc{ImpactPlate}~\citep{yuLearningFlexibleBody2023} and~\textsc{DeformingPlate}~\citep{pfaffLearningMeshBasedSimulation2020} considered in this work. The column \textit{Nonlinearity} distinguishes three different types: geometry (Geo), material (Mat), and boundary conditions (BC).  } 
{
    \centering
    \small
    \begin{tabularx}{\textwidth}{ l   m{1.6cm} m{1.8cm} m{1.5cm} m{1.5cm} m{1.0cm} m{0.8cm} }

            \hline
            \textbf{Datasets} & \textbf{Dynamic} & \textbf{Nonlinearity}& \textbf{avg. \#~Nodes} & \textbf{Mesh Type}  & \textbf{Steps T} & \textbf{Dim}\\
            \hline 
            
            \textsc{BendingBeam} &      Quasi-Static &  Geo &               744 &    Triangles &           400 & 2D    \\
            \textsc{BendingBeamLarge} &      Quasi-Static &  Geo &          8897 &    Triangles &           100 & 2D    \\

            \textsc{ImpactPlate} &      Dynamic &      Geo, BC &            2208 &         Triangles &                      52 & 2D     \\

            \textsc{DeformingPlate} &    Quasi-Static &   Geo, BC, Mat & 1271 &        Tetrahedrons &         400 & 3D    \\

            \hline
            
    \end{tabularx}\par

}
\label{table:datasets}
\end{table}

%% file: 05_inputs/figures/bb_force_bc.tex
\begin{figure}[!htb]
    \centering
    \includegraphics[width=\textwidth]{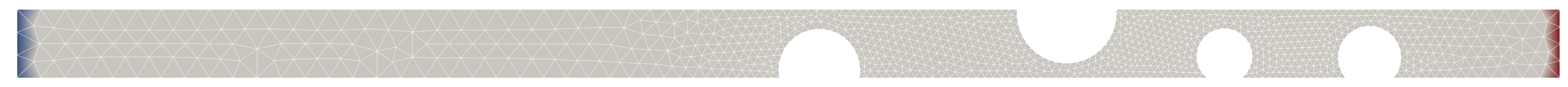}
    \caption{Visualization of the different node types on the~\textsc{BendingBeam} experiment. Blue nodes mark handle boundaries with fixed node positions. The red color indicates nodes where the force boundary condition is applied. Those locally applied forces in combination with the global, geometry-dependent part stiffness mainly determine the global deformation.}
    \label{fig:bb_force_bc}
\end{figure}

%% file: 02_appendix/0X_setup.tex
\section{Setup}\label{apx:setup}

\textbf{Hardware and Compute.}
We train all models on a single NVIDIA A100 GPU with a maximum training time of~$48$ hours, while most models required approximately~$40$ hours.
In total, we trained $11$ models across $3$ datasets, each on $5$ seeds:~\ac{robin},~\ac{mgn},~\ac{hcmt},~\ac{bsms}, as well as $7$ ablations of~\ac{robin}.
That amounts to $40~\text{hours} \times 5 \times 3 \times 11 = 6600~\text{hours}$ training time.
Furthermore, we trained two models on~\textsc{BendingBeamLarge}: a pre-trained~\ac{robin} model and a model from scratch, each on $5$ seeds for approximately~$40$ hours, yielding a total duration of $40~\text{hours} \times 5 \times 2 = 400 ~\text{hours}$.
For each training, we required a comparable amount of time for development and hyperparameter tuning. 
Additionally, we conducted inference experiments to measure the inference speed for $4$ models (\ac{robin},~\ac{mgn},~\ac{hcmt}, and~\ac{bsms}) and $7$~\ac{robin} variants across $5$ different seeds.
We run inference experiments on $3$ datasets, with each experiment taking about $1$ hour on average.
In total we obtain a runtime of $1~\text{hour} \times 5 \times 3 \times 11 = 165 ~\text{hours}$ for the inference experiments.

\textbf{Training.} We implement~\ac{robin} in \textit{PyTorch}~\citep{paszkePytorchImperativeStyle2019} and train it with ADAM~\citep{kingmaAdamMethodStochastic2017}.
We use an exponential learning rate decay, which decreases the learning rate from $1e-4$ to $1e-6$ over the training time, including $1000$ linearly increasing warm‑up steps.
We clip gradients such that their $L_2$-norm doesn't exceed $1$.
We train~\ac{robin} in~\textsc{BendingBeam} with $9$M samples and in~\textsc{ImpactPlate} with $6$M samples both with a batch size of $16$, resulting in $562\text{,}500$ and $375\text{,}000$ training iterations. 
In~\textsc{DeformingPlate} we reduce the batch size to $12$ and train for $300\text{,}000$ iterations with $3.6$M samples.

\textbf{Features.}
\Cref{table:features} provides an overview of the used input and output features for~\ac{robin}.
In addition to the default features, we extend the node embeddings of~\textsc{BendingBeam} with the force residual $\Delta f_{\text{BC}}$, which is defined by the boundary condition. 
In~\textsc{ImpactPlate}, we add the density $\rho_i$ and the Young's modulus $Y_i$ as node features. 
In~\textsc{DeformingPlate}, we add the scripted displacement residual $\Delta \mathbf{u}_{\text{BC}}$ of the actuator.
We normalize all input features based on the training dataset, setting them to have a zero mean and unit variance.
We add a small amount of training noise~\citep{sanchez-gonzalezLearningSimulateComplex2020,pfaffLearningMeshBasedSimulation2020} of $10^{-5}~\mathbf{\sigma}_{\mathbf{x}}$ to the node positions $\mathbf{x}_{i}^t$, where we scale the noise level with the standard deviation of the features $\mathbf{\sigma}_{\mathbf{x}}$.
For~\textsc{ImpactPlate} we noise the input history $\Delta \mathbf{u}^{t-1}_{i,0} = \mathbf{x}^{t}_{i} - \mathbf{x}^{t-1}_{i}$ with $ 10^{-3}~\sigma_{\mathbf{x}}$ to prevent overfitting on the history. 

\input{05_inputs/tables/features}

\textbf{Hierarchical Graph.} 
Since the relative motion of the components in the considered experiments is not too large, we define the contact edges based on the initial mesh configuration and keep them fixed to maintain a constant graph.
In~\textsc{DeformingPlate} we set the contact radius to $R=0.1$, connecting actuator nodes with plate nodes. 
In~\textsc{ImpactPlate} we connect ball nodes and plate nodes with a radius of $R=1.2$.
In all three experiments, we create $L=2$ coarse layers to obtain $3$ mesh levels. 

\textbf{\aclp{ampn}.} 
We use $3$ Pre- and $3$ Post-processing layers, $2$ Up- and $2$ Downsampling layers and $5$ Solving layers, which yields a total number of $15$ learnable layers.
We add a layer norm before each~\ac{mlp} and use two linear layers, a hidden size of $128$ and a~\ac{silu}~\citep{elfwingSigmoidweightedLinearUnits2018a} activation function. 
A max aggregation is used in all message passing layers. 

\textbf{\aclp{ddpm}.} 
We use $K=20$ denoising steps and a denoising stride of $m=5$ for~\ac{robin} by default. 
The $\beta$ variances of the~\ac{ddpm} scheduler are geometrically spaced for training and inference, starting from a minimum noise variance of $1e-4$ (for $\beta_1$) and going up to $1.0$ (for $\beta_K$).

\textbf{Metrics.} To compare the rollout accuracy, we follow~\citep{yuLearningFlexibleBody2023} and define the Root Mean Squared Euclidean distance error $\text{RMSE}= \sqrt{ 1/(N_i N_j) \sum_{i=1}^{N_i} \sum_{j=1}^{N_j} (\tilde{u}_{ij} - u_{ij})^2}$, where the prediction $\tilde{u}_{ij}$ and the ground truth ${u}_{ij}$ have $N_i$ nodes and $N_j$ features. 
We then calculate the mean over all time steps, the mean over the dataset and finally the mean $\mu$ and standard deviation $\sigma$ over the $5$ seeds.

%% file: 05_inputs/tables/features.tex
\newcolumntype{k}{>{\centering\arraybackslash\hsize=.2\hsize}X}
\newcolumntype{i}{>{\centering\arraybackslash\hsize=.175\hsize}X}
\newcolumntype{b}{>{\centering\arraybackslash\hsize=.3\hsize}X}
\begin{table}[H]
\caption{Node $\mathbf{k}_i$ and edge embeddings $\mathbf{e}_{ij}$ for the different datasets, depending on the node $\mathcal{V}$ and edge sets $\mathcal{E}$. } 
{
    \centering
    \small
    \begin{tabularx}{\textwidth}{ i   m{2.5cm} m{0.6cm} m{1.3cm} m{1.3cm} m{1.3cm} m{0.8cm} }

            \hline
            \textbf{Datasets} & \textbf{Inputs} $\mathcal{V}^0$  & \textbf{Inputs} $\mathcal{V}^{1:L}$& \textbf{Inputs} $\mathcal{E}^{0:L,\text{M}}$ & \textbf{Inputs} $\mathcal{E}^{0:L,\text{C}}$  & \textbf{Inputs} $\mathcal{E}^{0:L,\text{U/D}}$ & \textbf{Outputs} $\mathcal{V}^{0,\text{M}}$\\
            \hline 
            
            \textsc{BendingBeam} &      $\mathbf{n}_i$, $\Delta\mathbf{ u}^t_{i,k}$, $\Delta f_{\text{BC}}$ &  $\mathbf{n}_i$ &       ${\mathbf{x}}^t_{ij}$,$|{\mathbf{x}}^t_{ij}|$, ${\mathbf{x}}^0_{ij}$,$|{\mathbf{x}}^0_{ij}|$ &    ${\mathbf{x}}^t_{ij}$,$|{\mathbf{x}}^t_{ij}|$ &           ${\mathbf{x}}^t_{ij}$,$|{\mathbf{x}}^t_{ij}|$, ${\mathbf{x}}^0_{ij}$,$|{\mathbf{x}}^0_{ij}|$ & ${\mathbf{v}}_{i,\theta}(\Delta\mathbf{ u}^t_{i,k})$    \\

            \textsc{ImpactPlate} &      $\mathbf{n}_i$, $\Delta \mathbf{u}^t_{i,k}$, $\Delta\mathbf{ u}^{t-1}_{i,0}$, $\rho_i$, $Y_i$ &                           $\mathbf{n}_i$ &        ${\mathbf{x}}^t_{ij}$,$|{\mathbf{x}}^t_{ij}|$, ${\mathbf{x}}^0_{ij}$,$|{\mathbf{x}}^0_{ij}|$ &         ${\mathbf{x}}^t_{ij}$,$|{\mathbf{x}}^t_{ij}|$ &                      ${\mathbf{x}}^t_{ij}$,$|{\mathbf{x}}^t_{ij}|$, ${\mathbf{x}}^0_{ij}$,$|{\mathbf{x}}^0_{ij}|$ & ${\mathbf{v}}_{i,\theta}(\Delta\mathbf{ u}^t_{i,k})$     \\

            \textsc{DeformingPlate} &    $\mathbf{n}_i$, $\Delta \mathbf{ u}^t_{i,k}$, $\Delta \mathbf{ u}_{\text{BC}}$ &   $\mathbf{n}_i$&      ${\mathbf{x}}^t_{ij}$,$|{\mathbf{x}}^t_{ij}|$, ${\mathbf{x}}^0_{ij}$,$|{\mathbf{x}}^0_{ij}|$ &        ${\mathbf{x}}^t_{ij}$,$|{\mathbf{x}}^t_{ij}|$ &         ${\mathbf{x}}^t_{ij}$,$|{\mathbf{x}}^t_{ij}|$, ${\mathbf{x}}^0_{ij}$,$|{\mathbf{x}}^0_{ij}|$ & ${\mathbf{v}}_{i,\theta}(\Delta\mathbf{ u}^t_{i,k})$    \\

            \hline
            
    \end{tabularx}\par

}
\label{table:features}
\end{table}

%% file: 02_appendix/0X_baselines.tex
\section{Baselines, Ablations and Variants}\label{apx:baselines}

\textbf{Baselines.} 
We use the official~\textit{TensorFlow}~\citep{abadiTensorFlowSystemLargeScale2016} implementation of the authors for the baselines
\ac{hcmt}\footnote{\url{https://github.com/yuyudeep/hcmt/tree/main}}~\citep{yuLearningFlexibleBody2023} and
\ac{mgn}\footnote{\url{https://github.com/google-deepmind/deepmind-research/tree/master/meshgraphnets}}~\citep{pfaffLearningMeshBasedSimulation2020}.
We use ADAM~\citep{kingmaAdamMethodStochastic2017} for training~\ac{hcmt} and~\ac{mgn} with an exponential learning rate decay from $1e{-}4$ to $1e{-}6$, a batch size of $1$ and a hidden size of $128$.
We use $15$ message‑passing steps for~\ac{mgn} on all datasets, as well as a total of $15$~\ac{hmt} and~\ac{cmt} layers for~\ac{hcmt}.

On \textsc{BendingBeam}, we train \ac{hcmt} for $4$M training iterations with a training noise~\citep{sanchez-gonzalezLearningSimulateComplex2020,pfaffLearningMeshBasedSimulation2020} of $0.001$.
We maximize the receptive field and set the number of mesh levels to $5$, the maximum at which at least five nodes remain available across all meshes in the dataset, required for Delaunay remeshing~\citep{leeTwoAlgorithmsConstructing1980}.
This results in $9$~\acp{hmt}.
Since~\textsc{BendingBeam} is a contact-free task, we replaced the dual-branch~\ac{cmt} by $6$ single-branch Mesh Transformer layers that only attend to mesh edges instead of mesh and contact edges. 
We use the same architecture and hyperparameter for~\ac{hcmt} on~\textsc{ImpactPlate} and~\textsc{DeformingPlate} as proposed by the authors~\citep{yuLearningFlexibleBody2023}, and train it for $3$M steps and $2$M steps, respectively.

We follow the authors' implementation and add world edges to the mesh graph of~\ac{mgn} instead of contact edges to increase the receptive field of the non-hierarchical architecture.
\ac{mgn} is trained for $3$M iterations on~\textsc{BendingBeam} and uses a training noise of $0.001$ with a world edge radius of $R=0.13$.  
On~\textsc{ImpactPlate} we train~\ac{mgn} for $3$M steps and use a world edge radius of $R=0.03$ and a training noise of $0.003$.
We train~\ac{mgn} on~\textsc{DeformingPlate} for $1.5$M steps and use the authors' proposed settings~\citep{pfaffLearningMeshBasedSimulation2020}.
To prevent out-of-memory errors in edge cases on~\textsc{DeformingPlate}, we restrict the number of world edges to $200\text{,}000$ by selecting those with the smallest node distances.

For~\ac{bsms}~\citep{caoEfficientLearningMeshBased2023}, we use the official~\emph{PyTorch}~\citep{paszkePytorchImperativeStyle2019} implementation\footnote{\url{https://github.com/Eydcao/BSMS-GNN/tree/main}} of the authors.
We follow the authors and use the maximum number of hierarchy levels possible to maximize the receptive field. 
We train \ac{bsms} on~\textsc{BendingBeam} with a batch size of $12$ and $6$M training samples, i.e., $500$K training iterations. 
We use a training noise of $0.001$ and $5$ mesh levels. 
On~\textsc{ImpactPlate} we train~\ac{bsms} with a batch size of $8$ and $5$M samples ($625$K training iterations), a training noise of $0.003$, a contact radius of $R=0.4$, and $7$ hierarchy levels.
We train~\ac{bsms} on~\textsc{DeformingPlate} with a batch size of $8$ and $6$M samples ($750$K training iterations), a training noise of $0.003$, a contact radius of $R=0.03$, and $6$ hierarchy levels.

\textbf{Ablations.} 
The~\emph{10 diffusion steps} and~\emph{5 diffusion steps} ablation use the same settings as~\ac{robin}, despite the reduced number of diffusion steps $K$.
For the \emph{w/o hierarchy} ablation we use the fine mesh graph $\mathcal{G}^0$ instead of the hierarchical graph $\mathcal{G}^{0:L}$, replace our~\ac{ampn} by a single~\ac{intramps} with $15$ learnable message passing steps and remove the positional level encoding. 
In addition, we follow~\ac{mgn} and replace contact edges with world edges to increase the receptive field. 
To stay within the training budget, we reduce the number of training samples to ${1.2}$M for~\textsc{BendingBeam} and to ${8}$M for~\textsc{ImpactPlate}. 
For~\textsc{DeformingPlate} we reduce the batch size to $1$ and the training samples to ${0.6}$M and also restrict the number of world edges to $200\text{,}000$ as for~\ac{mgn}.
The \emph{w/o diffusion} ablation trains the~\ac{ampn} with an~\ac{mse} loss to predict directly the displacement residual $\Delta \mathbf{u}^t_{i,0}$ and uses a one‑step autoregressive rollout, such as~\ac{hcmt},~\ac{mgn}, and~\ac{bsms}. We use the same training noise settings as the baselines to stabilize the rollouts. 
The \emph{w/o shared layer} ablation uses a total number of $15$ non-shared learnable message passing layers distributed as follows: $1$ Pre-Processing and $1$ Post-Processing layer per level, $1$ Up- and $1$ Downsampling layer between each level, and $5$ Solving layers. 
The faster predictions allow an increase in the number of training samples to ${11}$M for~\textsc{BendingBeam}, to ${8}$M for~\textsc{ImpactPlate}, and to ${4.6}$M for~\textsc{DeformingPlate}. 
For the \emph{HCMT model} ablation, we replace the~\acp{ampn} with~\acp{hcmt}. 
More specifically, we use the same mesh hierarchy and the same model architecture as~\acp{hcmt}. 
Everything else remains the same in~\ac{robin}, including~\acp{ddpm} and~\ac{robi}.


%% file: 02_appendix/0X_results.tex
\section{Results}\label{apx:results}

\paragraph{\ac{amg}-based mesh coarsening.}
Root-node~\ac{amg} coarsening~\citep{vanekAlgebraicMultigridSmoothed1996b} preserves mesh geometry and connectivity (cf. \Cref{fig:bb_mesh_comparison}).
The coarse mesh remains well-aligned with thin geometrical features, and smoothed transfer operators yield wider, algebraically informed receptive fields compared to bi-stride pooling. 
This fidelity is crucial for predicting geometrically nonlinear deformations.

\input{05_inputs/figures/bb_mesh_comparison}

\paragraph{Inference speed.}
Each point in~\Cref{fig:ap_pareto} corresponds to a rollout setting of \ac{robin} on~\textsc{ImpactPlate} and~\textsc{DeformingPlate}. 
As for~\textsc{BendingBeam} in~\Cref{sec:results}, decreasing the truncation step reduces wall time while a truncation step of $k_\text{tr}{=}2$ is already sufficient to obtain a higher accuracy than all baselines across all datasets (cf.~\Cref{table:results_table}).
The \ac{robin} default $(1/20)$ is significantly faster than conventional diffusion inference and is even slightly more accurate on \textsc{ImpactPlate}, which we attribute to anchoring low-frequency components and reducing drift.

\input{05_inputs/figures/pareto_appendix}

\paragraph{Baselines.}
\Cref{fig:res_qual_ip_comparison} and \Cref{fig:res_qual_dp_comparison} visualize the rollout displacement and von Mises stress prediction of~\ac{robin},~\ac{hcmt},~\ac{mgn}, and \ac{bsms} on~\textsc{ImpactPlate} and~\textsc{DeformingPlate}, respectively.

\input{05_inputs/figures/ip_comparison}

\input{05_inputs/figures/dp_comparison}

\textbf{Diffusion truncation.} 
The rollouts in~\Cref{fig:res_truncation_qual} across truncation steps $k_\text{tr}$ illustrate coarse-to-fine frequency behavior of~\ac{robin}.
Early steps capture the global, low-frequency shape, whereas additional steps sharpen high-frequency details, e.g., high stresses in thin geometrical features.
Even with early truncation,~\ac{robin} maintains coherent global modes.
Longer schedules primarily refine local features and minimize the accumulation of high-frequency errors, while retaining the global deformation pattern.

\Cref{fig:res_truncation_ip_dp} visualize the rollout displacement RMSE and displacement gradient RMSE of different \ac{robin} variants on~\textsc{DeformingPlate} and~\textsc{ImpactPlate}. 
As for~\textsc{BendingBeam}, early denoising steps are critical for reducing global displacement error. Later diffusion steps focus on high-frequency solution components.
On \textsc{ImpactPlate}, the gradient RMSE shows a modest increase while \ac{robin} attains the lowest displacement RMSE overall. 
We hypothesize that partially denoising states (small $m$) stabilizes low-frequency components and reduces drift.
In contrast, fully denoising (larger $m$) suppresses short-term, high-frequency error accumulation. 
However, this trade-off does not affect other datasets with significantly longer rollouts.

\input{05_inputs/figures/truncation_qual}
\input{05_inputs/figures/truncation_ip_dp}


\paragraph{Quantitative results.}
\Cref{table:results_table} lists the quantitative results of all experiments considered in this work and used for the visualizations.

\input{05_inputs/tables/results_baselines}

%% file: 05_inputs/figures/bb_mesh_comparison.tex
\begin{figure}[!htb]
    \centering
    \includegraphics[width=\textwidth,trim={70pt, 150pt, 75pt, 150pt},  clip,]{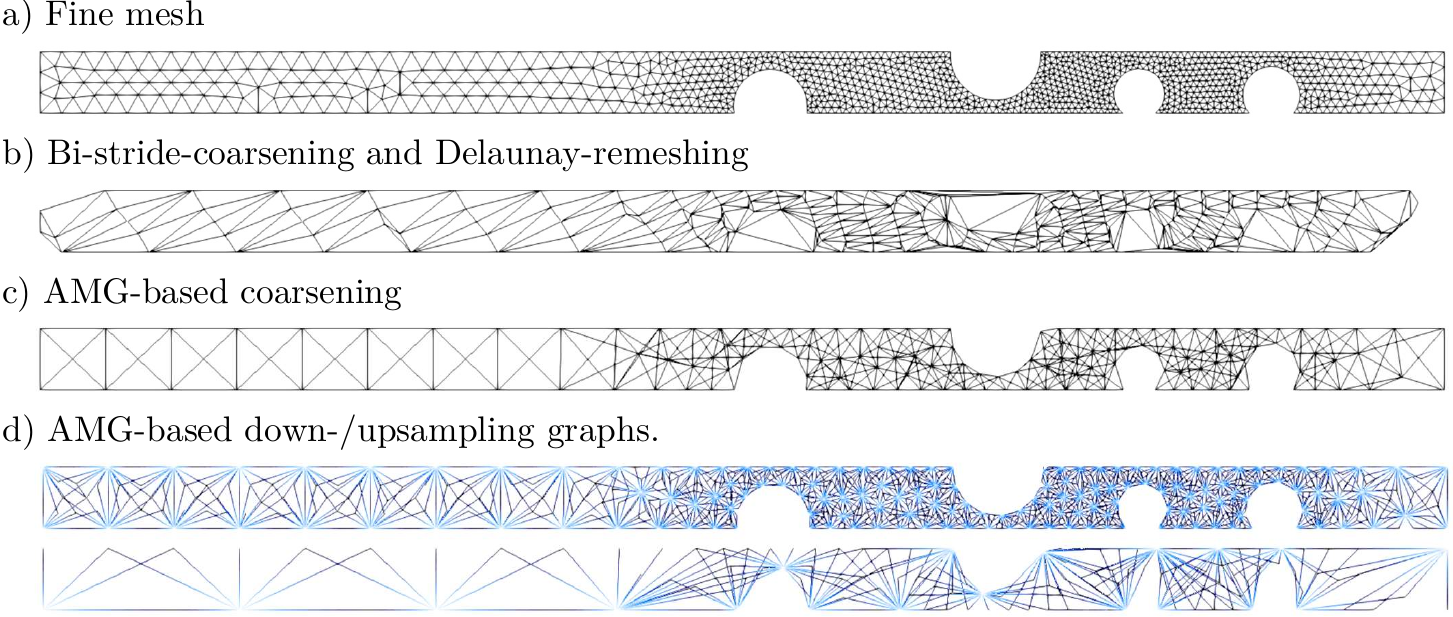}
    \caption{
    Comparison of~\ac{amg}-based mesh coarsening to~\ac{bsdl}. 
    \textbf{a)} the original, fine mesh.
    \textbf{b)} the mesh after two \ac{bsdl} coarsening steps.
    \textbf{c)} the mesh after one \ac{amg} coarsening step. This mesh has approximately as many nodes as the mesh in \textbf{b)}.
    \textbf{d)} up- and downsampling edges from level $0$ to $1$ (top) and level $1$ to $2$ (bottom). Bright blue indicates coarse nodes of level $1$ (top) or $2$ (bottom), respectively.
    }
    \label{fig:bb_mesh_comparison}
\end{figure}

%% file: 05_inputs/figures/pareto_appendix.tex

\begin{figure}[htbp] 
    \centering 

    \begin{subfigure}[b]{0.48\textwidth} 
        \centering 
        \includegraphics[width=\textwidth,trim={5pt, 00pt, 0pt 10pt},  clip,]{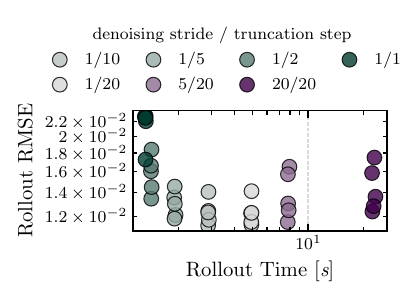}
        \caption{\textsc{ImpactPlate}. }
        \label{fig:baseline_comparison}
    \end{subfigure}
    \hfill 
    \begin{subfigure}[b]{0.48\textwidth} 
        \centering 
        \includegraphics[width=\textwidth,trim={5pt, 00pt, 0pt 10pt},  clip,]{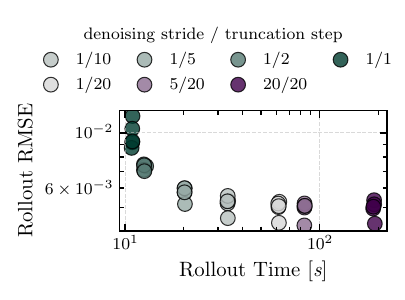}
        \caption{\textsc{DeformingPlate}.}
        \label{fig:pareto}
    \end{subfigure}

    \caption{RMSE rollout error and inference time of different inference variants of~\ac{robin} on a)~\textsc{ImpactPlate} and b)~\textsc{DeformingPlate}.~\ac{robin}, i.e., the variant (1/20), is most accurate on~\textsc{ImpactPlate} and on par on~\textsc{DeformingPlate} with the slower variants (5/20) and conventional inference (20/20). Reducing the truncation step $k_\text{tr}$ increases speed while decreasing accuracy. The one step variant (1/1) achieves the largest speed-up on~\textsc{DeformingPlate}, but also loses the most accuracy there.}
    \label{fig:ap_pareto}
\end{figure}

%% file: 05_inputs/figures/ip_comparison.tex
\begin{figure}[H]
    \vspace{-2.0cm}
    \centering
    \includegraphics[width=\textwidth,trim={20pt, 10pt, 00pt, 00pt},  clip,]{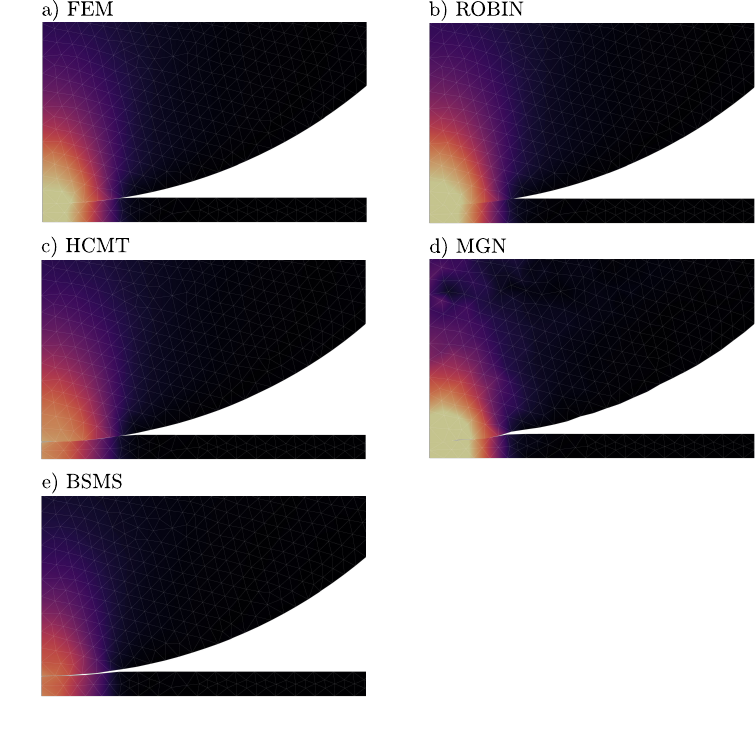}
    \caption{Comparison of the rollout deformation prediction and von Mises stress prediction (color, yellow is high) on~\textsc{ImpactPlate} to the ground truth of the~\ac{fem}.~\ac{robin} most accurately resolves the deformation at the contact surface and the resulting stress. The deformation prediction of~\ac{hcmt} and \ac{bsms} are close to the~\ac{fem} prediction, though stress is underestimated.~\ac{mgn} predicts accurately the global modes but exhibits local disturbances. }
    \label{fig:res_qual_ip_comparison}
    \vspace{5.2cm}
\end{figure}

%% file: 05_inputs/figures/dp_comparison.tex
\begin{figure}[H]
    \centering
    \includegraphics[width=\textwidth,trim={10pt, 10pt, 10pt, 10pt},  clip,]{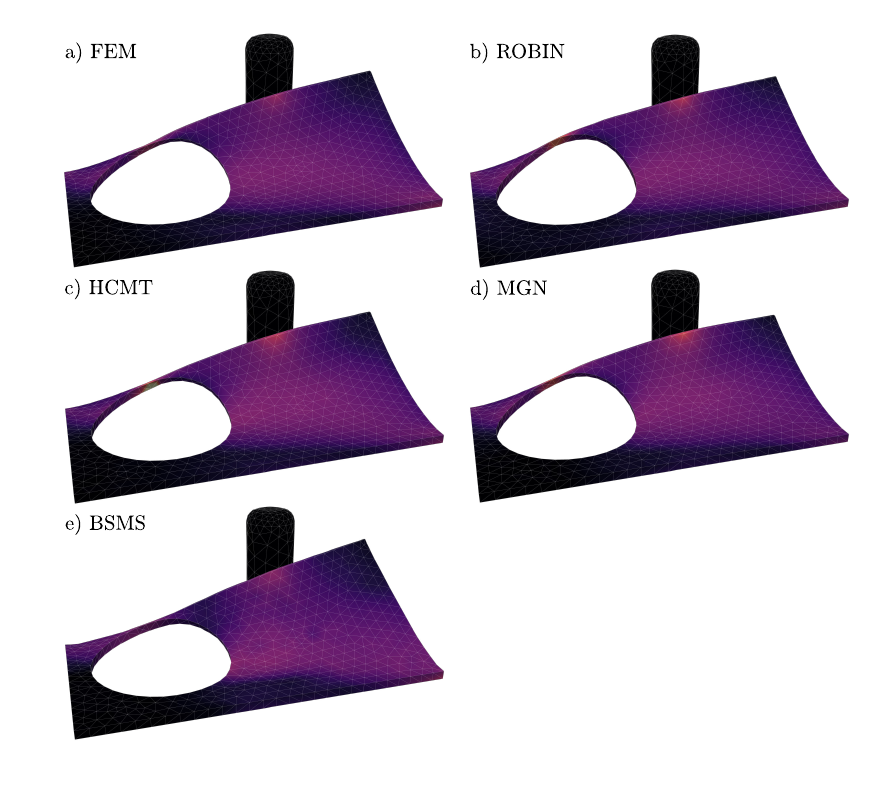}
    \caption{Rollout deformation and von Mises stress prediction (color, yellow is high) on~\textsc{DeformingPlate} of~\ac{robin}, the baselines and the~\ac{fem}. All models accurately reproduce the part deformation.~\ac{hcmt} slightly overestimates and~\ac{bsms} slightly underestimates the stress at the thin wall between the hole and the boundary. }
    \label{fig:res_qual_dp_comparison}
\end{figure}

%% file: 05_inputs/figures/truncation_qual.tex
\begin{figure}[!htb]
    \centering
    \includegraphics[width=\textwidth,trim={0pt, 0pt, 0pt 0pt},  ]{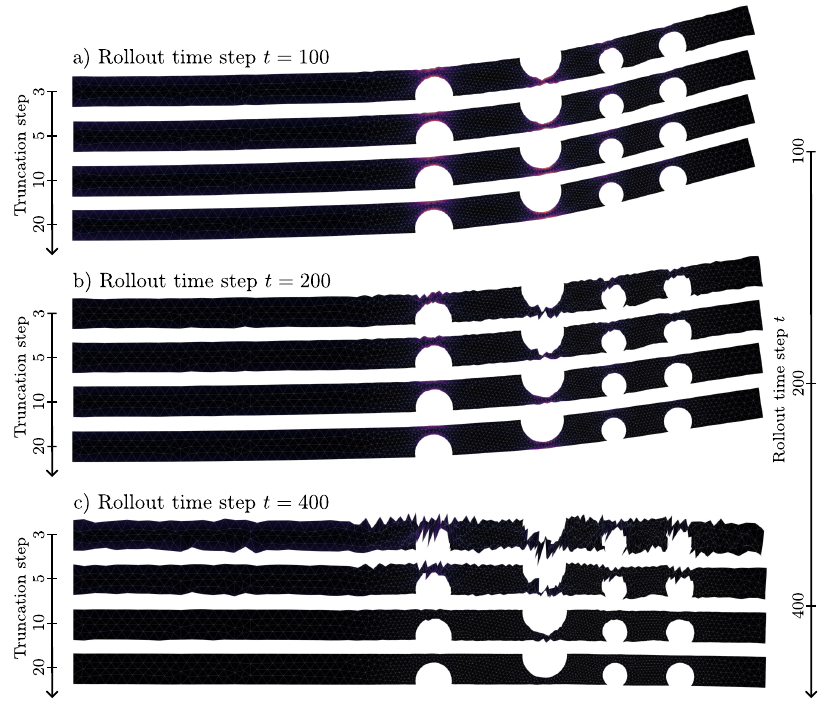}
    \caption{Effect of diffusion truncation on rollouts. 
    Figures a), b), and c) show predicted deformations and von Mises stress prediction (color, yellow is high) at $t=100$, $t=200$, and $t=400$, respectively. In each figure, rows correspond to truncation steps $k_\text{tr} = 3,5,10,20$.
    Using a low truncation step $k_\text{tr}$ increases inference speed and enables robust predictions of the global deformation modes. However, it also causes local mesh degradation due to the accumulation of high-frequency errors, as observed in~\ac{mse} trained one step models~\citep{lippePDERefinerAchievingAccurate2023}. 
    }
    \label{fig:res_truncation_qual}
    \vspace{-0.3cm}
\end{figure}

%% file: 05_inputs/figures/truncation_ip_dp.tex
\begin{figure}[htbp] 
    \centering 

    \begin{subfigure}[b]{0.48\textwidth} 
        \centering 
        \includegraphics[width=\textwidth,trim={5pt, 00pt, 0pt 10pt},  clip,]{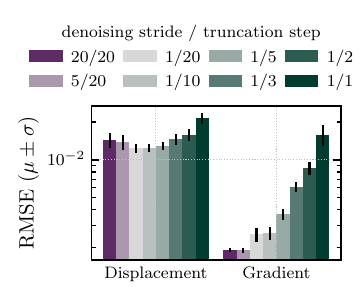}
        \caption{\textsc{ImpactPlate}. }
        \label{fig:res_truncation_ip}
    \end{subfigure}
    \hfill 
    \begin{subfigure}[b]{0.48\textwidth} 
        \centering 
        \includegraphics[width=\textwidth,trim={5pt, 00pt, 0pt 10pt},  clip,]{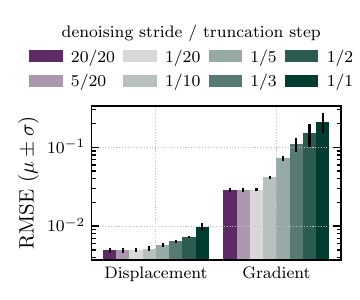}
        \caption{\textsc{DeformingPlate}.}
        \label{fig:res_truncation_dp}
    \end{subfigure}

    \caption{Comparison of the rollout displacement RMSE and displacement gradient RMSE for different~\ac{robin} setting (denoising stride $m$ / truncation step $k_\text{tr}$) on a)  \textsc{ImpactPlate} and b) \textsc{DeformingPlate}. 
    Early diffusion steps substantially reduce the displacement RMSE and later steps the local displacement gradient RMSE. 
    The fast inference setting (1/20) remains accurate and yields slightly lower displacement RMSE on~\textsc{ImpactPlate}, but also slightly higher gradient RMSE.
    }

    \label{fig:res_truncation_ip_dp}
\end{figure}

%% file: 05_inputs/tables/results_baselines.tex
\newcolumntype{k}{>{\centering\arraybackslash\hsize=0.8\hsize}X}
\newcolumntype{b}{>{\centering\arraybackslash\hsize=.85\hsize}X}
\begin{table}[h]
\caption{Quantitative results on \textsc{BendingBeam}, \textsc{ImpactPlate}, \textsc{DeformingPlate}, and the large-mesh dataset \textsc{BendingBeamLarge}. 
We report displacement RMSE [$10^{-3}$], gradient RMSE of the displacement field [$10^{-3}$], and rollout wall-clock time [s]. 
Values are mean~$\pm$~std over the test set across 5 seeds (solver timings excluded). 
The row \emph{Numerical solving} gives the average runtime of the high-fidelity solver (maximum in parentheses). 
\emph{Baselines} contrasts different models, \emph{Variants} sweep the~\ac{robi} settings (denoising stride / truncation step) and additionally report Gradient RMSE, \emph{Ablations} remove~\ac{robin} components, and \emph{Generalization to large meshes} evaluates the upscaling capabilities of~\ac{robin}.}

{
    \centering
    \scriptsize
    \begin{tabularx}{\textwidth}{ b  k k k k k k }

            \hline
             & \multicolumn{2}{c}{\textbf{\textsc{BendingBeam}}}  & \multicolumn{2}{c}{\textbf{\textsc{ImpactPlate}}} & \multicolumn{2}{c}{\textbf{\textsc{DeformingPlate}}} \\
            \hline 
            
            \textbf{Numerical solving} & \multicolumn{2}{c}{Time [s]}              &      \multicolumn{2}{c}{Time [s]}                  &    \multicolumn{2}{c}{Time [s]}  \\

            \hline 
            &     \multicolumn{2}{c}{46.2 (max. 186.1)}           & \multicolumn{2}{c}{742.6~\citep{yuLearningFlexibleBody2023}} & \multicolumn{2}{c}{1157.2~\citep{pfaffLearningMeshBasedSimulation2020}} \\

            \hline   
            \hline

            \textbf{Baselines} & & & & & & \\  

            \hline 
                                    & RMSE [$10^{-3}$]          & Time [s]              &  RMSE [$10^{-3}$]     & Time [s]                  & RMSE [$10^{-3}$]      & Time [s]\\

            \hline
            
            ROBIN              &  29.00 $\pm$ 1.00         & 15.03 $\pm$ 0.04      & 12.33 $\pm$ 0.96      & 4.94 $\pm$ 0.01           & 4.98 $\pm$ 0.33       & 61.60 $\pm$ 0.29 \\  
            HCMT               &   121.53 $\pm$ 1.87       & 25.82 $\pm$ 0.28      & 19.57 $\pm$ 0.38      & 4.33 $\pm$ 0.05           & 8.04 $\pm$ 0.13       & 31.57 $\pm$ 0.14 \\     
            MGN                &   189.52 $\pm$ 70.87      & 21.40 $\pm$ 0.15      & 54.07 $\pm$ 5.88      & 2.93 $\pm$ 0.03           & 8.76 $\pm$ 0.29       & 24.42 $\pm$ 0.22 \\    
            BSMS               &   141.98 $\pm$ 7.96       & 4.70 $\pm$ 0.04       & 63.52 $\pm$ 32.36     & 2.16 $\pm$ 0.01           & 13.59 $\pm$ 5.21      & 13.28 $\pm$ 0.07   \\                   
             
            \hline              
            \hline 
            
            \textbf{Variants (denoising stride / truncation step)} & & & & & & \\  
            \hline
                                    & RMSE [$10^{-3}$]          & Time [s]              &  RMSE [$10^{-3}$]     & Time [s]                  & RMSE [$10^{-3}$]      & Time [s]\\  
            \hline 

            20/20                   &  28.90 $\pm$ 1.91         & 162.43 $\pm$ 0.89     & 14.43 $\pm$ 1.93      & 22.61 $\pm$ 0.36          & 4.96 $\pm$ 0.36       & 190.10 $\pm$ 1.33 \\
            5/20                    &  28.99 $\pm$ 1.93         & 43.39 $\pm$ 0.52      & 13.87 $\pm$ 1.90      & 7.83 $\pm$ 0.06           & 4.92 $\pm$ 0.35      & 83.61 $\pm$ 0.16 \\
            1/20                    &  29.00 $\pm$ 1.00         & 15.03 $\pm$ 0.04      & 12.33 $\pm$ 0.96      & 4.94 $\pm$ 0.01           & 4.98 $\pm$ 0.33       & 61.60 $\pm$ 0.29 \\
            1/10                    &  30.35 $\pm$ 2.41         & 11.03 $\pm$ 0.10      & 12.38 $\pm$ 0.92      & 2.90  $\pm$ 0.01          & 5.18 $\pm$ 0.35       & 33.71 $\pm$ 0.10  \\
            1/5                     &  35.81 $\pm$ 3.30         & 9.88 $\pm$ 0.19       & 13.01 $\pm$ 0.98      & 1.91 $\pm$ 0.01           & 5.73 $\pm$ 0.30       & 20.24 $\pm$ 0.04  \\
            1/3                     &  42.83 $\pm$ 4.54         & 9.57 $\pm$ 0.11       & 14.58 $\pm$ 1.40      & 1.55 $\pm$ 0.01           & 6.42 $\pm$ 0.23       & 14.85 $\pm$ 0.11  \\
            1/2                     &  48.94 $\pm$ 7.25         & 9.47 $\pm$ 0.11       & 15.79 $\pm$ 1.71      & 1.42 $\pm$ 0.00           & 7.26 $\pm$ 0.18       & 12.59 $\pm$ 0.13  \\
            1/1                     &  64.23 $\pm$ 11.86        & 9.49 $\pm$ 0.12       & 21.46 $\pm$ 2.12      & 1.32 $\pm$ 0.01           & 10.79 $\pm$ 0.52      & 7.11 $\pm$ 0.04  \\
   
            \hline 
                                    & \multicolumn{2}{c}{Gradient RMSE [$10^{-3}$] }& \multicolumn{2}{c}{Gradient RMSE [$10^{-3}$]  }                      & \multicolumn{2}{c}{Gradient RMSE [$10^{-3}$]} \\

            \hline

            20/20                   & \multicolumn{2}{c}{12.75 $\pm$ 0.70          }&  \multicolumn{2}{c}{1.92 $\pm$ 0.07            } & \multicolumn{2}{c}{28.98 $\pm$ 1.56 } \\
            5/20                    & \multicolumn{2}{c}{12.74 $\pm$ 0.65          }&  \multicolumn{2}{c}{1.89 $\pm$ 0.08            } & \multicolumn{2}{c}{28.83 $\pm$ 1.57 } \\
            1/20                    & \multicolumn{2}{c}{12.77 $\pm$ 0.72          }&  \multicolumn{2}{c}{2.55 $\pm$ 0.33            } & \multicolumn{2}{c}{29.10 $\pm$ 1.50 } \\
            1/10                    & \multicolumn{2}{c}{37.58 $\pm$ 4.71          }&  \multicolumn{2}{c}{2.61 $\pm$ 0.30            } & \multicolumn{2}{c}{41.59 $\pm$ 1.76 }  \\
            1/5                     & \multicolumn{2}{c}{94.00 $\pm$ 20.90         }&  \multicolumn{2}{c}{3.68 $\pm$ 0.37            } & \multicolumn{2}{c}{73.13 $\pm$ 5.64 }  \\
            1/3                     & \multicolumn{2}{c}{147.97 $\pm$ 65.13        }&  \multicolumn{2}{c}{6.07 $\pm$ 0.56            } & \multicolumn{2}{c}{110.96 $\pm$ 22.33 } \\
            1/2                     & \multicolumn{2}{c}{200.44 $\pm$ 128.29       }&  \multicolumn{2}{c}{8.61 $\pm$ 1.02            } & \multicolumn{2}{c}{150.33 $\pm$ 48.48 } \\
            1/1                     & \multicolumn{2}{c}{324.05 $\pm$ 285.65       }&  \multicolumn{2}{c}{23.41 $\pm$ 4.85           } & \multicolumn{2}{c}{157.30 $\pm$ 8.34 } \\
            
            \hline   
            \hline 
            
            \textbf{Ablations} & & & & & & \\  

            \hline 
                                    & \multicolumn{2}{c}{RMSE [$10^{-3}$]          } &  \multicolumn{2}{c}{RMSE [$10^{-3}$]     }& \multicolumn{2}{c}{RMSE [$10^{-3}$]  } \\
            \hline 
            
            ROBIN              & \multicolumn{2}{c}{29.00 $\pm$ 1.00          } & \multicolumn{2}{c}{12.33 $\pm$ 0.96      }& \multicolumn{2}{c}{4.98 $\pm$ 0.33       }            \\  
            10 diff. steps        & \multicolumn{2}{c}{33.28  $\pm$ 2.43         } & \multicolumn{2}{c}{14.58 $\pm$ 1.99      }& \multicolumn{2}{c}{5.86  $\pm$ 0.61      }            \\  
            5 diff. steps           & \multicolumn{2}{c}{41.33  $\pm$ 6.27         } & \multicolumn{2}{c}{15.60 $\pm$ 1.74      }& \multicolumn{2}{c}{6.40  $\pm$ 0.56      }            \\  
            w/o shared layer      & \multicolumn{2}{c}{66.05   $\pm$ 7.36        } & \multicolumn{2}{c}{12.88 $\pm$ 1.58      }& \multicolumn{2}{c}{5.65  $\pm$ 0.26      }            \\  
            State prediction          & \multicolumn{2}{c}{596.38   $\pm$ 70.93      } & \multicolumn{2}{c}{130.50 $\pm$ 30.80    }& \multicolumn{2}{c}{137.61  $\pm$ 30.54   }             \\  
            w/o hierarchy   & \multicolumn{2}{c}{122.38   $\pm$ 1.52       } & \multicolumn{2}{c}{57.02  $\pm$ 3.48     }& \multicolumn{2}{c}{12.04  $\pm$ 0.24     }            \\  
            w/o diffusion   & \multicolumn{2}{c}{43.83 $\pm$ 1.16          } & \multicolumn{2}{c}{99.47  $\pm$ 47.30    }& \multicolumn{2}{c}{9.78  $\pm$ 2.12      }            \\  
            HCMT model           & \multicolumn{2}{c}{111.05   $\pm$ 1.15       } & \multicolumn{2}{c}{29.18 $\pm$ 1.43      }& \multicolumn{2}{c}{14.86 $\pm$ 0.42      }            \\  

            \hline   
                        & & & & & & \\ 
            \hline 
            
            & \multicolumn{2}{c}{\textbf{\textsc{BendingBeamLarge}}}  & & & & \\
            \hline 
            \textbf{Numerical solving} & \multicolumn{2}{c}{Time [s]}              &                        &  & &   \\

            \hline 
            
                    &     \multicolumn{2}{c}{108.30 (max. 4248.01)}                      & & & & \\ 
            
            \hline 
            \hline 
            
            \textbf{Generalization to large meshes} & & & & & & \\ 

            \hline
             & RMSE [$10^{-3}$]   & Time [s]  & & & & \\
            \hline

            Pre-trained             & 77.59 $\pm$ 4.97 & 30.94 $\pm$ 0.63 & & & & \\ 
            
            From scratch            & 215.50 $\pm$ 12.14 & 30.95 $\pm$ 0.58 & & & & \\ 
            
            \hline

    \end{tabularx}\par

}
\label{table:results_table}
\end{table}

%% file: 02_appendix/0X_neurips_checklist.tex
\newpage

\section*{NeurIPS Paper Checklist}

\begin{enumerate}

\item {\bf Claims}
    \item[] Question: Do the main claims made in the abstract and introduction accurately reflect the paper's contributions and scope?
    \item[] Answer: \answerYes{} 
    \item[] Justification: The claims in the abstract and introduction are fully supported by our method section, as well as in the qualitative and quantitative results in the experiments section. The appendix provides more detailed results where required.
    \item[] Guidelines:
    \begin{itemize}
        \item The answer NA means that the abstract and introduction do not include the claims made in the paper.
        \item The abstract and/or introduction should clearly state the claims made, including the contributions made in the paper and important assumptions and limitations. A No or NA answer to this question will not be perceived well by the reviewers. 
        \item The claims made should match theoretical and experimental results, and reflect how much the results can be expected to generalize to other settings. 
        \item It is fine to include aspirational goals as motivation as long as it is clear that these goals are not attained by the paper. 
    \end{itemize}

\item {\bf Limitations}
    \item[] Question: Does the paper discuss the limitations of the work performed by the authors?
    \item[] Answer: \answerYes{} 
    \item[] Justification: The conclusion discusses current limitations of the approach, including the scope of the paper and assumptions made.
    \item[] Guidelines:
    \begin{itemize}
        \item The answer NA means that the paper has no limitation while the answer No means that the paper has limitations, but those are not discussed in the paper. 
        \item The authors are encouraged to create a separate "Limitations" section in their paper.
        \item The paper should point out any strong assumptions and how robust the results are to violations of these assumptions (e.g., independence assumptions, noiseless settings, model well-specification, asymptotic approximations only holding locally). The authors should reflect on how these assumptions might be violated in practice and what the implications would be.
        \item The authors should reflect on the scope of the claims made, e.g., if the approach was only tested on a few datasets or with a few runs. In general, empirical results often depend on implicit assumptions, which should be articulated.
        \item The authors should reflect on the factors that influence the performance of the approach. For example, a facial recognition algorithm may perform poorly when image resolution is low or images are taken in low lighting. Or a speech-to-text system might not be used reliably to provide closed captions for online lectures because it fails to handle technical jargon.
        \item The authors should discuss the computational efficiency of the proposed algorithms and how they scale with dataset size.
        \item If applicable, the authors should discuss possible limitations of their approach to address problems of privacy and fairness.
        \item While the authors might fear that complete honesty about limitations might be used by reviewers as grounds for rejection, a worse outcome might be that reviewers discover limitations that aren't acknowledged in the paper. The authors should use their best judgment and recognize that individual actions in favor of transparency play an important role in developing norms that preserve the integrity of the community. Reviewers will be specifically instructed to not penalize honesty concerning limitations.
    \end{itemize}

\item {\bf Theory assumptions and proofs}
    \item[] Question: For each theoretical result, does the paper provide the full set of assumptions and a complete (and correct) proof?
    \item[] Answer: \answerNA{} 
    \item[] Justification: The paper does not introduce any new theoretical results.
    \item[] Guidelines:
    \begin{itemize}
        \item The answer NA means that the paper does not include theoretical results. 
        \item All the theorems, formulas, and proofs in the paper should be numbered and cross-referenced.
        \item All assumptions should be clearly stated or referenced in the statement of any theorems.
        \item The proofs can either appear in the main paper or the supplemental material, but if they appear in the supplemental material, the authors are encouraged to provide a short proof sketch to provide intuition. 
        \item Inversely, any informal proof provided in the core of the paper should be complemented by formal proofs provided in appendix or supplemental material.
        \item Theorems and Lemmas that the proof relies upon should be properly referenced. 
    \end{itemize}

    \item {\bf Experimental result reproducibility}
    \item[] Question: Does the paper fully disclose all the information needed to reproduce the main experimental results of the paper to the extent that it affects the main claims and/or conclusions of the paper (regardless of whether the code and data are provided or not)?
    \item[] Answer: \answerYes{} 
    \item[] Justification: We detail our method in Section~\ref{sec:robin}, and provide additional information in the appendix where required. We detail our experimental setup and datasets used in the experiments section.
    \item[] Guidelines:
    \begin{itemize}
        \item The answer NA means that the paper does not include experiments.
        \item If the paper includes experiments, a No answer to this question will not be perceived well by the reviewers: Making the paper reproducible is important, regardless of whether the code and data are provided or not.
        \item If the contribution is a dataset and/or model, the authors should describe the steps taken to make their results reproducible or verifiable. 
        \item Depending on the contribution, reproducibility can be accomplished in various ways. For example, if the contribution is a novel architecture, describing the architecture fully might suffice, or if the contribution is a specific model and empirical evaluation, it may be necessary to either make it possible for others to replicate the model with the same dataset, or provide access to the model. In general. releasing code and data is often one good way to accomplish this, but reproducibility can also be provided via detailed instructions for how to replicate the results, access to a hosted model (e.g., in the case of a large language model), releasing of a model checkpoint, or other means that are appropriate to the research performed.
        \item While NeurIPS does not require releasing code, the conference does require all submissions to provide some reasonable avenue for reproducibility, which may depend on the nature of the contribution. For example
        \begin{enumerate}
            \item If the contribution is primarily a new algorithm, the paper should make it clear how to reproduce that algorithm.
            \item If the contribution is primarily a new model architecture, the paper should describe the architecture clearly and fully.
            \item If the contribution is a new model (e.g., a large language model), then there should either be a way to access this model for reproducing the results or a way to reproduce the model (e.g., with an open-source dataset or instructions for how to construct the dataset).
            \item We recognize that reproducibility may be tricky in some cases, in which case authors are welcome to describe the particular way they provide for reproducibility. In the case of closed-source models, it may be that access to the model is limited in some way (e.g., to registered users), but it should be possible for other researchers to have some path to reproducing or verifying the results.
        \end{enumerate}
    \end{itemize}

\item {\bf Open access to data and code}
    \item[] Question: Does the paper provide open access to the data and code, with sufficient instructions to faithfully reproduce the main experimental results, as described in supplemental material?
    \item[] Answer: \answerNo{} 
    \item[] Justification: We do not provide data and code at the time of submission, but will open-source both after acceptance.
    \item[] Guidelines:
    \begin{itemize}
        \item The answer NA means that paper does not include experiments requiring code.
        \item Please see the NeurIPS code and data submission guidelines (\url{https://nips.cc/public/guides/CodeSubmissionPolicy}) for more details.
        \item While we encourage the release of code and data, we understand that this might not be possible, so “No” is an acceptable answer. Papers cannot be rejected simply for not including code, unless this is central to the contribution (e.g., for a new open-source benchmark).
        \item The instructions should contain the exact command and environment needed to run to reproduce the results. See the NeurIPS code and data submission guidelines (\url{https://nips.cc/public/guides/CodeSubmissionPolicy}) for more details.
        \item The authors should provide instructions on data access and preparation, including how to access the raw data, preprocessed data, intermediate data, and generated data, etc.
        \item The authors should provide scripts to reproduce all experimental results for the new proposed method and baselines. If only a subset of experiments are reproducible, they should state which ones are omitted from the script and why.
        \item At submission time, to preserve anonymity, the authors should release anonymized versions (if applicable).
        \item Providing as much information as possible in supplemental material (appended to the paper) is recommended, but including URLs to data and code is permitted.
    \end{itemize}

\item {\bf Experimental setting/details}
    \item[] Question: Does the paper specify all the training and test details (e.g., data splits, hyperparameters, how they were chosen, type of optimizer, etc.) necessary to understand the results?
    \item[] Answer: \answerYes{} 
    \item[] Justification: We specify high-level training and test details in Section~\ref{sec:experiments}, and provide further details in Appendices~\ref{apx:baselines},~\ref{apx:datasets} and~\ref{apx:setup}.
    \item[] Guidelines:
    \begin{itemize}
        \item The answer NA means that the paper does not include experiments.
        \item The experimental setting should be presented in the core of the paper to a level of detail that is necessary to appreciate the results and make sense of them.
        \item The full details can be provided either with the code, in appendix, or as supplemental material.
    \end{itemize}

\item {\bf Experiment statistical significance}
    \item[] Question: Does the paper report error bars suitably and correctly defined or other appropriate information about the statistical significance of the experiments?
    \item[] Answer: \answerYes{} 
    \item[] Justification: We report mean and standard deviation across five seeds for all main experiments. When evaluating the runtime-error tradeoff, we directly report information for all seeds without aggregation, explicitly showing the performance difference between runs.
    \item[] Guidelines:
    \begin{itemize}
        \item The answer NA means that the paper does not include experiments.
        \item The authors should answer "Yes" if the results are accompanied by error bars, confidence intervals, or statistical significance tests, at least for the experiments that support the main claims of the paper.
        \item The factors of variability that the error bars are capturing should be clearly stated (for example, train/test split, initialization, random drawing of some parameter, or overall run with given experimental conditions).
        \item The method for calculating the error bars should be explained (closed form formula, call to a library function, bootstrap, etc.)
        \item The assumptions made should be given (e.g., Normally distributed errors).
        \item It should be clear whether the error bar is the standard deviation or the standard error of the mean.
        \item It is OK to report 1-sigma error bars, but one should state it. The authors should preferably report a 2-sigma error bar than state that they have a 96\% CI, if the hypothesis of Normality of errors is not verified.
        \item For asymmetric distributions, the authors should be careful not to show in tables or figures symmetric error bars that would yield results that are out of range (e.g. negative error rates).
        \item If error bars are reported in tables or plots, The authors should explain in the text how they were calculated and reference the corresponding figures or tables in the text.
    \end{itemize}

\item {\bf Experiments compute resources}
    \item[] Question: For each experiment, does the paper provide sufficient information on the computer resources (type of compute workers, memory, time of execution) needed to reproduce the experiments?
    \item[] Answer: \answerYes{} 
    \item[] Justification: We provide information on the compute resources for all experiments in Appendix~\ref{apx:setup}.
    \item[] Guidelines:
    \begin{itemize}
        \item The answer NA means that the paper does not include experiments.
        \item The paper should indicate the type of compute workers CPU or GPU, internal cluster, or cloud provider, including relevant memory and storage.
        \item The paper should provide the amount of compute required for each of the individual experimental runs as well as estimate the total compute. 
        \item The paper should disclose whether the full research project required more compute than the experiments reported in the paper (e.g., preliminary or failed experiments that didn't make it into the paper). 
    \end{itemize}
    
\item {\bf Code of ethics}
    \item[] Question: Does the research conducted in the paper conform, in every respect, with the NeurIPS Code of Ethics \url{https://neurips.cc/public/EthicsGuidelines}?
    \item[] Answer: \answerYes{} 
    \item[] Justification: We have read the NeurIPS Code of Ethics. We made sure that our research complies to the Code of Ethics in every respect.
    \item[] Guidelines:
    \begin{itemize}
        \item The answer NA means that the authors have not reviewed the NeurIPS Code of Ethics.
        \item If the authors answer No, they should explain the special circumstances that require a deviation from the Code of Ethics.
        \item The authors should make sure to preserve anonymity (e.g., if there is a special consideration due to laws or regulations in their jurisdiction).
    \end{itemize}

\item {\bf Broader impacts}
    \item[] Question: Does the paper discuss both potential positive societal impacts and negative societal impacts of the work performed?
    \item[] Answer: \answerYes{} 
    \item[] Justification: We include a discussion on broader impact in Appendix~\ref{apx:broader_impact}.
    \item[] Guidelines:
    \begin{itemize}
        \item The answer NA means that there is no societal impact of the work performed.
        \item If the authors answer NA or No, they should explain why their work has no societal impact or why the paper does not address societal impact.
        \item Examples of negative societal impacts include potential malicious or unintended uses (e.g., disinformation, generating fake profiles, surveillance), fairness considerations (e.g., deployment of technologies that could make decisions that unfairly impact specific groups), privacy considerations, and security considerations.
        \item The conference expects that many papers will be foundational research and not tied to particular applications, let alone deployments. However, if there is a direct path to any negative applications, the authors should point it out. For example, it is legitimate to point out that an improvement in the quality of generative models could be used to generate deepfakes for disinformation. On the other hand, it is not needed to point out that a generic algorithm for optimizing neural networks could enable people to train models that generate Deepfakes faster.
        \item The authors should consider possible harms that could arise when the technology is being used as intended and functioning correctly, harms that could arise when the technology is being used as intended but gives incorrect results, and harms following from (intentional or unintentional) misuse of the technology.
        \item If there are negative societal impacts, the authors could also discuss possible mitigation strategies (e.g., gated release of models, providing defenses in addition to attacks, mechanisms for monitoring misuse, mechanisms to monitor how a system learns from feedback over time, improving the efficiency and accessibility of ML).
    \end{itemize}
    
\item {\bf Safeguards}
    \item[] Question: Does the paper describe safeguards that have been put in place for responsible release of data or models that have a high risk for misuse (e.g., pretrained language models, image generators, or scraped datasets)?
    \item[] Answer: \answerNA{} 
    \item[] Justification: We do not use pretrained language models, image generators or similar high-risk models in our approach, and do not scrape datasets.
    We still discuss potential cases for miss-use of our learned simulator in Appendix~\ref{apx:broader_impact}.
    \item[] Guidelines:
    \begin{itemize}
        \item The answer NA means that the paper poses no such risks.
        \item Released models that have a high risk for misuse or dual-use should be released with necessary safeguards to allow for controlled use of the model, for example by requiring that users adhere to usage guidelines or restrictions to access the model or implementing safety filters. 
        \item Datasets that have been scraped from the Internet could pose safety risks. The authors should describe how they avoided releasing unsafe images.
        \item We recognize that providing effective safeguards is challenging, and many papers do not require this, but we encourage authors to take this into account and make a best faith effort.
    \end{itemize}

\item {\bf Licenses for existing assets}
    \item[] Question: Are the creators or original owners of assets (e.g., code, data, models), used in the paper, properly credited and are the license and terms of use explicitly mentioned and properly respected?
    \item[] Answer: \answerYes{} 
    \item[] Justification: We use two publicly available datasets, namely ImpactPlate and DeformingPlate. We credit the original papers of both explicitly.
    \item[] Guidelines:
    \begin{itemize}
        \item The answer NA means that the paper does not use existing assets.
        \item The authors should cite the original paper that produced the code package or dataset.
        \item The authors should state which version of the asset is used and, if possible, include a URL.
        \item The name of the license (e.g., CC-BY 4.0) should be included for each asset.
        \item For scraped data from a particular source (e.g., website), the copyright and terms of service of that source should be provided.
        \item If assets are released, the license, copyright information, and terms of use in the package should be provided. For popular datasets, \url{paperswithcode.com/datasets} has curated licenses for some datasets. Their licensing guide can help determine the license of a dataset.
        \item For existing datasets that are re-packaged, both the original license and the license of the derived asset (if it has changed) should be provided.
        \item If this information is not available online, the authors are encouraged to reach out to the asset's creators.
    \end{itemize}

\item {\bf New assets}
    \item[] Question: Are new assets introduced in the paper well documented and is the documentation provided alongside the assets?
    \item[] Answer: \answerNA{} 
    \item[] Justification: At time of submission, we do not release new assets. After acceptance, we will open-source our BendingBeam dataset, and provide appropriate documentation alongside it.
    \item[] Guidelines:
    \begin{itemize}
        \item The answer NA means that the paper does not release new assets.
        \item Researchers should communicate the details of the dataset/code/model as part of their submissions via structured templates. This includes details about training, license, limitations, etc. 
        \item The paper should discuss whether and how consent was obtained from people whose asset is used.
        \item At submission time, remember to anonymize your assets (if applicable). You can either create an anonymized URL or include an anonymized zip file.
    \end{itemize}

\item {\bf Crowdsourcing and research with human subjects}
    \item[] Question: For crowdsourcing experiments and research with human subjects, does the paper include the full text of instructions given to participants and screenshots, if applicable, as well as details about compensation (if any)? 
    \item[] Answer: \answerNA{} 
    \item[] Justification: We do not involve crowdsourcing nor research with human subjects.
    \item[] Guidelines:
    \begin{itemize}
        \item The answer NA means that the paper does not involve crowdsourcing nor research with human subjects.
        \item Including this information in the supplemental material is fine, but if the main contribution of the paper involves human subjects, then as much detail as possible should be included in the main paper. 
        \item According to the NeurIPS Code of Ethics, workers involved in data collection, curation, or other labor should be paid at least the minimum wage in the country of the data collector. 
    \end{itemize}

\item {\bf Institutional review board (IRB) approvals or equivalent for research with human subjects}
    \item[] Question: Does the paper describe potential risks incurred by study participants, whether such risks were disclosed to the subjects, and whether Institutional Review Board (IRB) approvals (or an equivalent approval/review based on the requirements of your country or institution) were obtained?
    \item[] Answer: \answerNA{} 
    \item[] Justification: The paper does not involve crowdsourcing nor research with human subjects.
    \item[] Guidelines:
    \begin{itemize}
        \item The answer NA means that the paper does not involve crowdsourcing nor research with human subjects.
        \item Depending on the country in which research is conducted, IRB approval (or equivalent) may be required for any human subjects research. If you obtained IRB approval, you should clearly state this in the paper. 
        \item We recognize that the procedures for this may vary significantly between institutions and locations, and we expect authors to adhere to the NeurIPS Code of Ethics and the guidelines for their institution. 
        \item For initial submissions, do not include any information that would break anonymity (if applicable), such as the institution conducting the review.
    \end{itemize}

\item {\bf Declaration of LLM usage}
    \item[] Question: Does the paper describe the usage of LLMs if it is an important, original, or non-standard component of the core methods in this research? Note that if the LLM is used only for writing, editing, or formatting purposes and does not impact the core methodology, scientific rigorousness, or originality of the research, declaration is not required.
    \item[] Answer: \answerNA{} 
    \item[] Justification: The core method development in this research does not involve LLMs as any important, original, or non-standard components.
    \item[] Guidelines:
    \begin{itemize}
        \item The answer NA means that the core method development in this research does not involve LLMs as any important, original, or non-standard components.
        \item Please refer to our LLM policy (\url{https://neurips.cc/Conferences/2025/LLM}) for what should or should not be described.
    \end{itemize}

\end{enumerate}